\documentclass[11pt]{article}

\usepackage[preprint]{acl}

\usepackage{times}
\usepackage{latexsym}
\usepackage{makecell}
\usepackage{enumitem}

\usepackage[T1]{fontenc}

\usepackage[utf8]{inputenc}

\usepackage{microtype}

\usepackage{inconsolata}

\usepackage{graphicx}

\usepackage{booktabs}
\usepackage{multirow}
\usepackage{amsmath}
\usepackage{amssymb}\usepackage{amsfonts}\usepackage{afterpage}

\newcommand{\BOS}{\texttt{<BOS>} }

\title{Garbage Attention in Large Language Models:\\
\texttt{<BOS>} Sink Heads and Sink-aware Pruning}

\author{%
Jaewon Sok$^{1}$ \quad
Jewon Yeom$^{2}$ \quad
Seonghyeon Park$^{3}$ \quad
Jeongjae Park$^{2}$ \quad
Taesup Kim$^{2,}$\thanks{\ \ Corresponding author.} \\
$^{1}$Department of Rural Systems Engineering, Seoul National University\\
$^{2}$Graduate School of Data Science, Seoul National University\\
$^{3}$Department of Aerospace Engineering, Seoul National University
}

\begin{document}
\maketitle

\begin{abstract}
Large Language Models (LLMs) are known to contain significant redundancy, yet a systematic explanation for why certain components, particularly in higher layers, are more redundant has remained elusive.
In this work, we identify the \BOS sink phenomenon as a key mechanism driving this layer-wise sensitivity.
We show that attention heads with high \BOS sink scores are strongly associated with functional redundancy: such heads, especially in deeper layers, contribute little to predictive performance and effectively serve as \emph{dumping grounds} for superfluous attention weights.
Leveraging this insight, we introduce a simple pruning strategy that removes high-\BOS sink heads.
Experiments on Gemma-3, Llama-3.1, and Qwen3 demonstrate that this approach identifies redundant transformer components more reliably than weight- and activation-based criteria in terms of downstream task retention, remaining close to dense baselines at low-to-moderate pruning ratios.
We further find that high-scoring sink heads sustain their focus on \BOS as context length grows. Overall, our results suggest that structural properties of attention offer a more direct basis for model compression than magnitude-based methods.
\end{abstract}

\begin{figure}[h]
    \centering
    \includegraphics[width=0.9\linewidth]{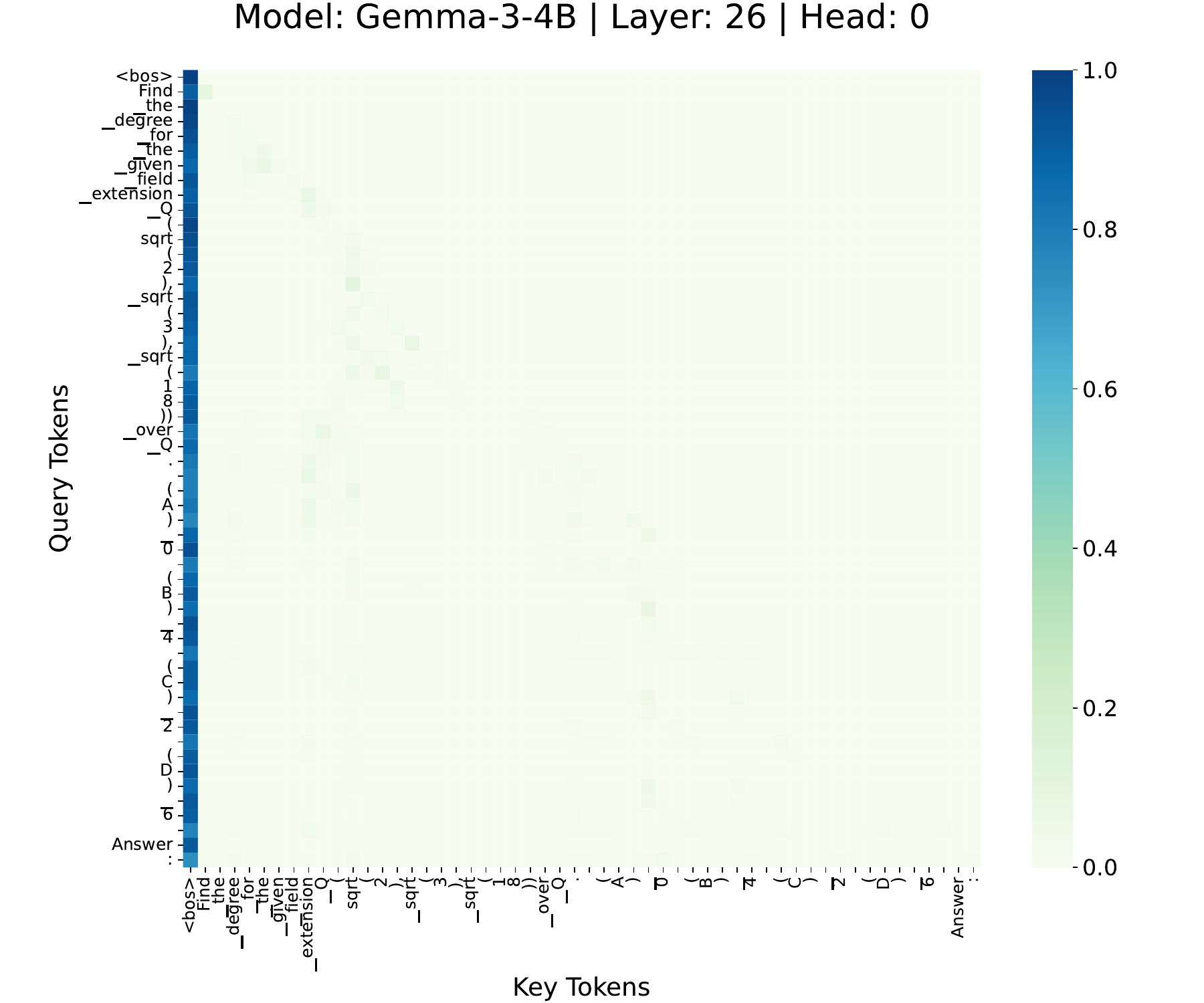}
    \caption{Attention weights of a representative \BOS sink head (L26, H0) in Gemma-3-4B. The heatmap reveals a stark attention sink pattern where query tokens disproportionately attend to the \BOS token during an MMLU task.}
    \label{fig:bos_sink_attention_map}
\end{figure}

\section{Introduction}

The advent of Transformers \citep{vaswani2017attention} has enabled Large Language Models (LLMs) to achieve unprecedented performance on various tasks \citep{brown2020language, zhao2023survey}. However, the massive parameter spaces of these models contain significant redundancy \citep{michel2019sixteen, fan2024not, he2024matters, kaushik2025universal}, serving as a primary bottleneck for efficient deployment. Consequently, while much effort has been devoted to model compression and pruning \citep{hoefler2021sparsity, 10.1162/tacl_a_00704, song2024sleb}, the underlying nature and origins of this redundancy remain relatively unexplored.

Previous research on the attention mechanism, the cornerstone of the Transformer architecture, has demonstrated that individual attention heads perform distinct functional roles \citep{clark2019does, voita-etal-2019-analyzing}. Notably, a phenomenon known as the \emph{attention sink} has been identified, where attention weights are disproportionately concentrated on specific tokens, particularly the Beginning-of-Sentence (\texttt{<BOS>}) token \citep{xiao2023efficient}. Further analysis reveals that this behavior is driven by specialized attention heads that exhibit a remarkably high propensity for such effects \citep{queipo2025attention}. The importance of attention sinks is evidenced by the substantial performance degradation observed when omitting the \BOS token at inference time in models trained with it \citep{barbero2025llms}.

In parallel, structural pruning studies have increasingly focused on the layer-wise sensitivity of LLMs to identify redundant components. A recurring observation in this line of work is that LLMs exhibit a non-uniform distribution of importance across their depth, with higher layers often being more amenable to pruning or skipping without catastrophic performance loss \citep{gromov2024unreasonable, zhang2024finercut, men2025shortgpt}. Recent studies further demonstrate that attention heads in higher layers can be effectively pruned, particularly when combined with rescaling to preserve representation stability \citep{liu2025high}.

Despite their importance during training, the functional role of attention sink heads at inference remains largely underexplored. Moreover, while prior pruning literature consistently reports redundancy in higher layers, clear criteria for identifying such layers are still lacking, and proposed explanations are often post hoc.

In this work, we make the following contributions:
\begin{itemize}[leftmargin=*]
    \item We identify \BOS sink heads (Figure~\ref{fig:bos_sink_attention_map}) as structurally specialized attention heads that act as \emph{dumping grounds} for superfluous attention weights, providing a mechanistic explanation for redundancy in LLMs.
    \item We show that the emergence and concentration of these heads in higher layers explain the layer-wise sensitivity consistently observed in prior pruning literature.
    \item We demonstrate that \BOS sink scores constitute a simple yet robust structural metric for identifying redundant heads and layers, enabling effective structured pruning while preserving predictive performance.
\end{itemize}


\section{Related Work}

\subsection{Attention Sink}

Attention sinks describe the tendency of LLMs to allocate disproportionate mass to early tokens, especially \BOS, as a stable outlet for the surplus attention probability \citep{xiao2023efficient, gu2024attention}. Previous work further suggests that this pattern is not merely incidental noise, but a learned allocation strategy that emerges during training to satisfy normalization constraints while preserving useful token-to-token interactions. Motivated by this account, \citet{miller2023attention} proposes adding a constant to the softmax denominator, which allows a head's attention weights to approach zero so that no token need serve as the outlet.

Recent analysis shows that the phenomenon is concentrated in specialized sink heads and aligned with highly redundant regions of the network \citep{queipo2025attention}, supporting the view that many of such heads operate with limited task-specific utility. At the same time, removing \BOS at inference can substantially degrade models trained to rely on it, indicating that structural anchoring and functional redundancy can coexist \citep{barbero2025llms}. This duality motivates a more granular perspective: the anchor token may remain globally important even when a subset of sink-specialized heads is locally expendable. Building on this tension, we ask whether sink heads, rather than the \BOS token itself, can be pruned safely and systematically.

\subsection{Pruning Methods}

Prior work reduces LLM cost through two broad paradigms: unstructured weight sparsification and structured removal of architectural units \citep{han2015learning, ma2023llm}. Unstructured methods target individual parameters and can achieve high sparsity, as exemplified by magnitude pruning, Wanda, SparseGPT, and Wanda++ \citep{sun2023simple, frantar2023sparsegpt, yang2025wanda++}. In contrast, structured methods remove entire components (e.g., heads) or full layers, often yielding clearer hardware compatibility and more direct latency gains \citep{an2023fluctuation, ashkboos2024slicegpt, men2025shortgpt}. Within structured pruning, this distinction is commonly framed as component-level versus layer-level pruning \citep{song2024sleb, yang2024laco}. Across these settings, multiple studies report that deeper blocks are often more prunable than earlier ones; however, selection criteria are typically derived from magnitude, saliency, similarity, or representation heuristics rather than from an explicit mechanism of functional redundancy \citep{men2025shortgpt, yang2025entropy}. Our work complements this line by introducing the \BOS sink score as a simple structural signal that links head-level redundancy to layer-wise pruning sensitivity.

\begin{figure*}[t]
    \centering
    \includegraphics[width=0.32\linewidth]{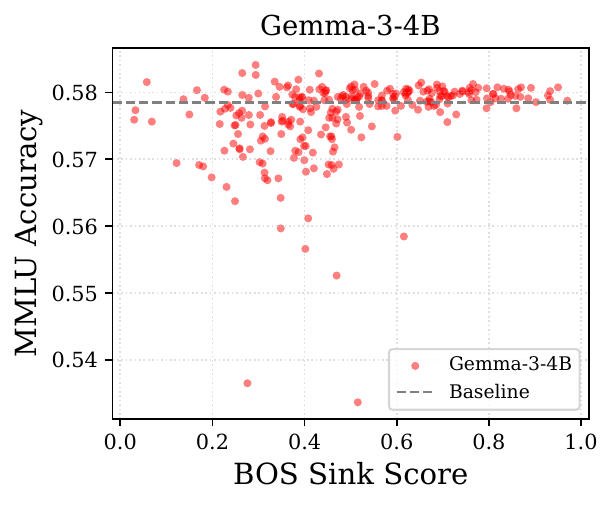}
    \includegraphics[width=0.32\linewidth]{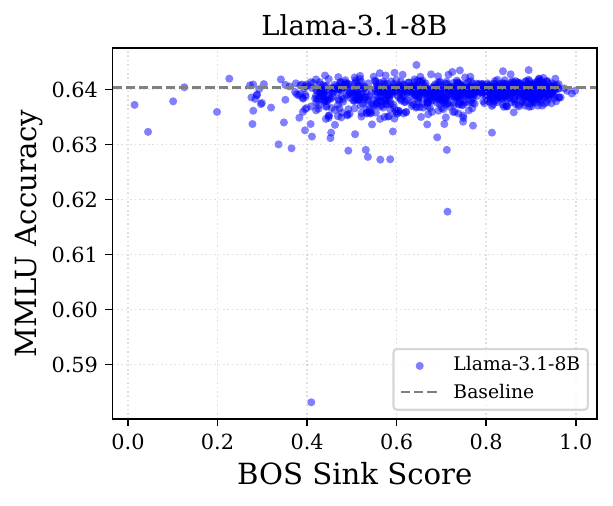}
    \includegraphics[width=0.32\linewidth]{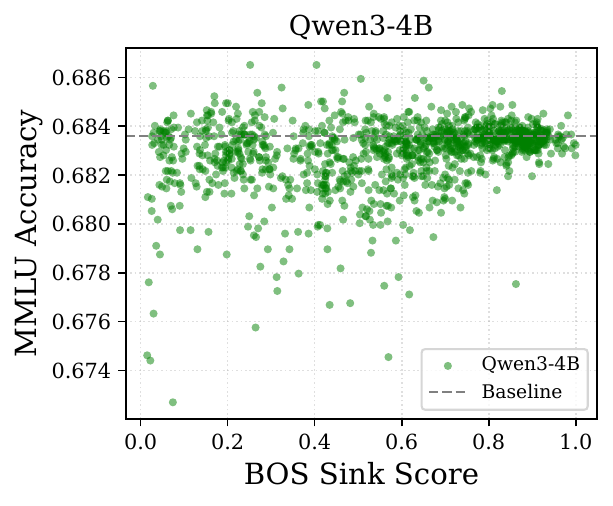}
    \caption{Single-head ablation on MMLU versus \BOS sink score for Gemma-3-4B, Llama-3.1-8B, and Qwen3-4B. Dots denote heads and dashed lines denote model baselines; high-score heads consistently have minimal impact when ablated.}
    \label{fig:head_ablation_comparison}
\end{figure*}

\begin{figure*}[t]
    \centering
    \includegraphics[width=0.32\linewidth]{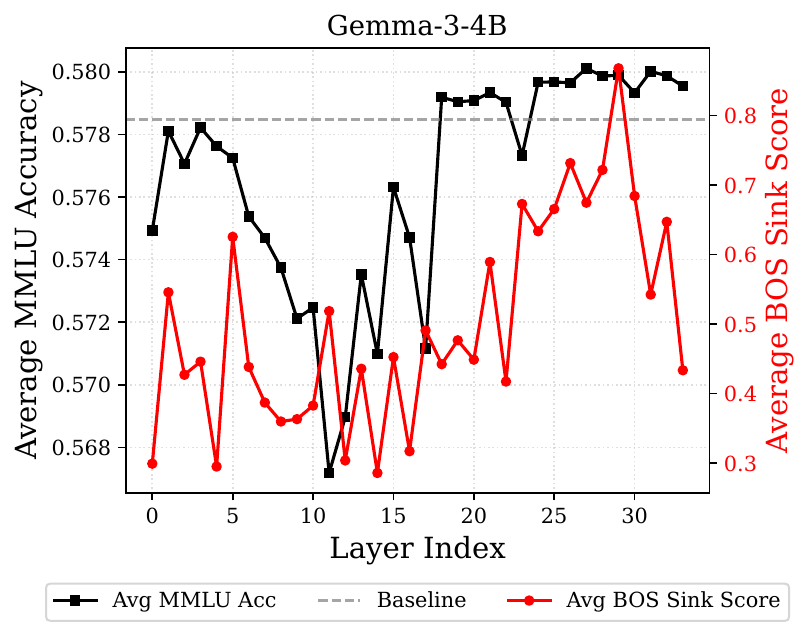}
    \includegraphics[width=0.32\linewidth]{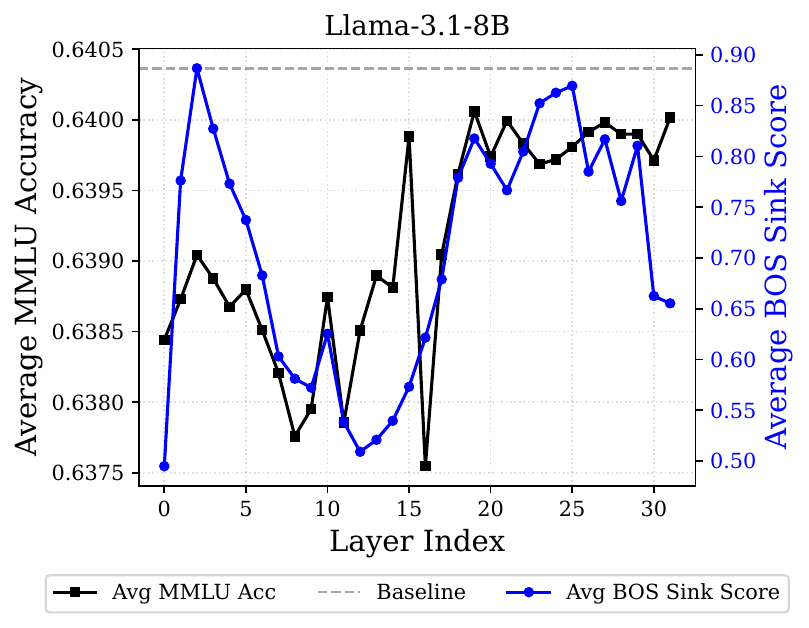}
    \includegraphics[width=0.32\linewidth]{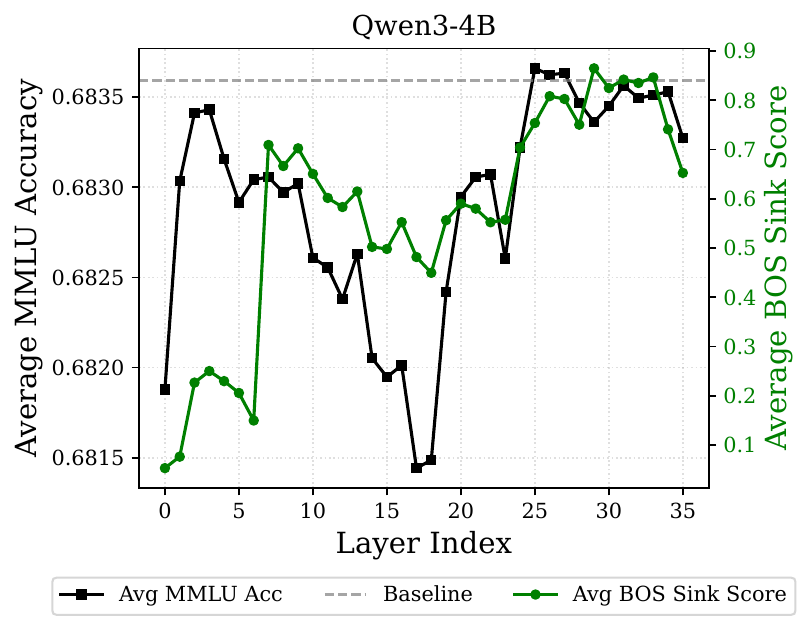}
    \caption{Relationship between \BOS sink score and single-head ablation effect on MMLU. Colored and black lines denote layer-mean sink score (right axis) and ablated MMLU accuracy (left axis), respectively. In deeper layers, higher sink scores coincide with lower functional necessity, indicating redundant heads.}
    \label{fig:layer_ablation_trend}
\end{figure*}

\section{Preliminary}

\subsection{Attention Sink Metrics}
To analyze how attention is allocated across different tokens, we adopt the metrics from \citet{gu2024attention}. Let $\alpha_{t, k}^{(\ell, h)}$ denote the attention weight from the query token at position $t$ to the key token at position $k$ in the $h$-th head of the $\ell$-th layer.

\paragraph{Attention Sink Score}
First, we use the attention sink score to quantify the degree to which a specific token position $k$ attracts attention weights within a head. For a sequence of length $T$, the score for token $k$ in head $h$ of layer $\ell$ is defined as the average attention weight it receives from all tokens in the sequence:
\begin{equation}
    \operatorname{sink-score}_k^{(\ell, h)} = \frac{1}{T} \sum_{t=0}^{T-1} \alpha_{t, k}^{(\ell, h)}
\end{equation}
A high score indicates that the token at position $k$ serves as a primary destination for attention weights within that specific head.

\paragraph{\BOS Sink Score}
Given that the \BOS token (at index $k=0$) often acts as a focal point for attention, we define the \BOS sink score to measure its influence at both the head and layer levels. The head-wise \BOS sink score $S_{\BOS}^{(\ell, h)}$ and the layer-wise \BOS sink score $S_{\BOS}^{(\ell)}$ are defined as follows:
\begin{equation}
\label{eq:bos_score}
\begin{aligned}
    S_{\BOS}^{(\ell, h)} &= \operatorname{sink-score}_0^{(\ell, h)} \\
    S_{\BOS}^{(\ell)} &= \frac{1}{H} \sum_{h=1}^{H} S_{\BOS}^{(\ell, h)}
\end{aligned}
\end{equation}
where $H$ is the total number of attention heads in layer $\ell$. The layer-wise \BOS sink score $S_{\BOS}^{(\ell)}$ serves as an aggregate metric to evaluate the dependence of layer $\ell$ on the \BOS token.

\subsection{Block Influence Score}
To identify redundant layers within LLMs, \citet{men2025shortgpt} leverages the Block Influence (BI) score within the ShortGPT method. This metric quantifies the degree of representation transformation performed by each layer by measuring the distance between its input and output hidden states.

\paragraph{Definition}
Let $X_{i,t} \in \mathbb{R}^d$ denote the $t$-th row of the hidden state matrix (corresponding to the $t$-th token) at the $i$-th layer. The BI score for the $i$-th layer is defined as:
\begin{equation}
    \text{BI}_i = 1 - \mathbb{E}_{X,t} \frac{X_{i,t}^T X_{i+1,t}}{\|X_{i,t}\|_2 \|X_{i+1,t}\|_2}
\end{equation}
where $\|\cdot\|_2$ denotes the $L_2$ norm, and the expectation $\mathbb{E}_{X,t}$ is taken over a dataset $X$ and token positions $t$.

\section{\BOS Sink as a Marker of Functional Redundancy}
\label{sec:bos_redundancy}

\paragraph{\BOS Sink Score Extraction}
To analyze the model's internal behavior, we calculate the \BOS sink score defined in Equation~\ref{eq:bos_score}. Specifically, we perform 0-shot inference on the MMLU dataset and extract attention weights directly from the attention maps during the initial decoding step.

\paragraph{Head-wise Ablation}
Figure~\ref{fig:head_ablation_comparison} plots single-head ablation impact on MMLU versus \BOS sink score for Llama-3.1-8B, Qwen3-4B, and Gemma-3-4B. Across all models, sensitivity separates into two regimes: high-\BOS heads are consistently near-baseline when removed, whereas low-\BOS heads induce larger variance and occasional accuracy drops. This low-score sensitivity is most pronounced in Gemma-3-4B, which we attribute to lower head density (272 heads) relative to Llama-3.1-8B (1,024) and Qwen3-4B (1,152), leaving less functional redundancy per head. Accordingly, high-\BOS redundancy appears architecture-invariant, while vulnerability to low-\BOS ablation increases as head density decreases.

\paragraph{Layer-wise Concentration in Deeper Blocks}
To characterize layer-wise redundancy, we measure the mean ablation impact of heads by layer. Figure~\ref{fig:layer_ablation_trend} shows that \BOS sink scores are concentrated in deeper layers. In these final blocks, higher mean \BOS scores coincide with near-zero average accuracy loss under single-head ablation, indicating strong functional redundancy.

Early layers exhibit model-specific behavior. In Qwen3-4B and Gemma-3-4B, ablation impact remains small despite low \BOS scores, whereas Llama-3.1-8B shows elevated early-layer \BOS scores with occasional accuracy decline. These results suggest that high \BOS scores are a reliable redundancy signal in deeper layers, while the role of \BOS in initial layers is architecture-dependent.

\section{Mechanistic Evidence: Sink Heads as Attention Dumping Grounds}

\paragraph{Operational Characteristics}
A \BOS sink head is defined by two primary traits: (i) a disproportionately high \BOS sink score $S_{\BOS}^{(\ell,h)}$, and (ii) near-zero functional sensitivity under ablation. As illustrated in Figure~\ref{fig:head_ablation_comparison}, heads in the high-\BOS regime across Gemma-3, Llama-3.1, and Qwen3 consistently show negligible impact on MMLU accuracy when removed. This behavior suggests that these heads do not transform or convey critical semantic information but instead serve as a stable outlet within the attention mechanism.

\paragraph{The Structural Uniqueness of \BOS and Redundancy}
A key enabler of this phenomenon is the structural uniqueness of \BOS. Unlike other tokens, it is fixed at position 0 and intentionally non-semantic, encoding neither lexical meaning nor contextual information. It thus provides a stable, content-free target for attention mass under softmax normalization, which forces weights to sum to one even when no semantically relevant target exists \citep{xiao2023efficient}. We hypothesize that as representations stabilize in deeper layers, some heads increasingly exploit this property and converge on \BOS as a non-semantic \emph{dumping ground}, whose invariant, context-independent status offers a consistent outlet for surplus attention probability. These heads then contribute to the residual stream in a near-invariant way, shifting from semantic processing to structural maintenance---a transition reflected in Appendix~\ref{sec:attention_patterns}.

\paragraph{Layer-wise Localization of Redundancy} The concentration of these dumping grounds in deeper blocks is driven by the stabilization of representations in higher layers. As the model reaches a consensus on the input's meaning, additional attention heads become functionally redundant, repurposed as numerical outlets to satisfy the softmax-induced requirement. This localized redundancy explains the minimal performance impact of pruning deeper layers. Consequently, high \BOS sink scores offer a robust proxy for identifying this layer-specific redundancy, providing a clear empirical basis for efficient structural pruning in the model's final stages.

\section{\texttt{<BOS>} Sink-based Structured Pruning}
\label{sec:pruning_method}
Based on the evidence of \BOS sink heads as redundancy repositories, we introduce a structured pruning strategy that employs the \BOS sink score to identify and eliminate redundant components.

\paragraph{Pruning Criterion}
We rank architectural components by their \BOS sink scores and prune those with the highest values. Attention heads are selected by $S_{\BOS}^{(\ell,h)}$, and transformer blocks are selected by the layer-wise score $S_{\BOS}^{(\ell)}$ (Equation~\ref{eq:bos_score}). To maintain stable representations, we exclude the first and last transformer blocks from pruning. The first block is preserved because it builds core lexical and positional features for later layers. The last block is preserved following ShortGPT \citep{men2025shortgpt}, which reports that removing its FFN can cause catastrophic performance degradation.

\paragraph{Implementation (Head Ablation)}
For head-wise pruning, we compute \BOS sink scores on MMLU (0-shot), rank heads by sink intensity, and select the highest-scoring heads as pruning candidates. We ablate each selected head by zeroing its slice in the output projection matrix ($W_O$). For a target head $h$ in layer $\ell$, we set $W_O^{(\ell,h)} \in \mathbb{R}^{d_\text{head} \times d_\text{model}}$ to zero:
\begin{equation}
    W_O^{(\ell,h)} \leftarrow \mathbf{0}
\end{equation}
This operation nullifies the contribution of head $h$ before residual aggregation, effectively removing its functional effect while preserving model structure. We do not apply post-pruning rescaling or layer-norm recalibration, so performance changes directly reflect the quality of \BOS sink score-based head selection.

\paragraph{Pruning Granularity}
Attention and FFN write to the residual stream as separate sublayers, yet we remove them together as
a block. In pre-normalized architectures the update each module writes into the residual stream
shrinks relative to the stream itself as depth increases \citep{men2025shortgpt}; at
those depths the hidden state is carried mostly by the residual path, so deleting a block's two
sublayers together perturbs it little more than deleting either one alone. The block is also the more
economical unit: on Gemma-3-4B a 0.125 ratio corresponds to 4 blocks but to 23 of the 34 attention
sublayers, and Appendix~\ref{sec:granularity} shows that removing the latter under the identical
\BOS ranking is far more damaging.

\begin{table*}[t]
\centering
\small
\resizebox{\linewidth}{!}{
\begin{tabular}{c | c | c | c | cccccccc | c}
\toprule
\textbf{Model} & \textbf{Strategy} & \makecell[c]{\textbf{Pruning}\\\textbf{Ratio}} & \textbf{Wikitext-2 $\downarrow$} & \textbf{ARC-c} & \textbf{ARC-e} & \textbf{BoolQ} & \textbf{HellaS.} & \textbf{MMLU} & \textbf{OBQA} & \textbf{PIQA} & \textbf{WG} & \textbf{Avg.} \\
\midrule\midrule
\multirow{4}{*}{Gemma-3-4B} & Dense & 0 & 10.21 & 0.533 & 0.816 & 0.791 & 0.580 & 0.597 & 0.352 & 0.788 & 0.725 & 0.648 \\
\cmidrule{2-13}
 & Mag-sp & \multirow{3}{*}{0.125} & 17.98 & 0.421 & 0.754 & 0.624 & 0.451 & 0.442 & 0.324 & 0.767 & 0.559 & 0.543 \\
 & Wanda-sp &  & 13.53 & 0.470 & 0.786 & 0.724 & 0.508 & 0.540 & 0.318 & 0.771 & 0.635 & 0.594 \\
 & \textbf{Ours (\texttt{<BOS>})} &  & $\mathbf{11.27}$ & $\mathbf{0.522}$ & $\mathbf{0.813}$ & $\mathbf{0.770}$ & $\mathbf{0.571}$ & $\mathbf{0.594}$ & $\mathbf{0.348}$ & $\mathbf{0.788}$ & $\mathbf{0.718}$ & $\mathbf{0.641}$ \\
\midrule
\multirow{4}{*}{Gemma-3-12B} & Dense & 0 & 10.184 & 0.650 & 0.872 & 0.854 & 0.645 & 0.741 & 0.354 & 0.807 & 0.803 & 0.716 \\
\cmidrule{2-13}
 & Mag-sp & \multirow{3}{*}{0.250} & 25.582 & 0.389 & 0.815 & 0.699 & 0.420 & 0.316 & 0.314 & 0.788 & 0.628 & 0.546 \\
 & Wanda-sp &  & 24.619 & 0.411 & 0.824 & 0.661 & 0.447 & 0.595 & 0.310 & 0.768 & 0.564 & 0.573 \\
 & \textbf{Ours (\texttt{<BOS>})} &  & $\mathbf{13.607}$ & $\mathbf{0.622}$ & $\mathbf{0.846}$ & $\mathbf{0.820}$ & $\mathbf{0.619}$ & $\mathbf{0.711}$ & $\mathbf{0.342}$ & $\mathbf{0.803}$ & $\mathbf{0.792}$ & $\mathbf{0.694}$ \\
\midrule
\multirow{4}{*}{Llama-3.1-8B} & Dense & 0 & 7.71 & 0.550 & 0.822 & 0.829 & 0.618 & 0.653 & 0.336 & 0.791 & 0.781 & 0.673 \\
\cmidrule{2-13}
 & Mag-sp & \multirow{3}{*}{0.250} & 315.21 & 0.225 & 0.612 & 0.414 & 0.321 & 0.229 & 0.220 & 0.695 & 0.507 & 0.403 \\
 & Wanda-sp &  & $\mathbf{10.75}$ & 0.410 & 0.796 & 0.760 & 0.487 & 0.344 & $\mathbf{0.330}$ & $\mathbf{0.779}$ & 0.668 & 0.572 \\
 & \textbf{Ours (\texttt{<BOS>})} &  & 20.55 & $\mathbf{0.521}$ & $\mathbf{0.810}$ & $\mathbf{0.787}$ & $\mathbf{0.589}$ & $\mathbf{0.524}$ & 0.316 & 0.778 & $\mathbf{0.722}$ & $\mathbf{0.631}$ \\
\midrule
\multirow{4}{*}{Qwen3-4B} & Dense & 0 & 17.31 & 0.584 & 0.804 & 0.851 & 0.532 & 0.701 & 0.294 & 0.753 & 0.672 & 0.649 \\
\cmidrule{2-13}
 & Mag-sp & \multirow{3}{*}{0.125} & 30.91 & 0.386 & 0.662 & 0.569 & 0.401 & 0.459 & 0.248 & 0.708 & 0.585 & 0.502 \\
 & Wanda-sp &  & $\mathbf{18.48}$ & 0.542 & 0.760 & 0.804 & 0.513 & 0.653 & $\mathbf{0.306}$ & 0.734 & 0.637 & 0.618 \\
 & \textbf{Ours (\texttt{<BOS>})} &  & 19.87 & $\mathbf{0.568}$ & $\mathbf{0.794}$ & $\mathbf{0.817}$ & $\mathbf{0.534}$ & $\mathbf{0.690}$ & 0.302 & $\mathbf{0.755}$ & $\mathbf{0.685}$ & $\mathbf{0.643}$ \\
\bottomrule
\end{tabular}
}
\caption{Performance comparison of component-level pruning strategies on Gemma-3-4B/12B, Llama-3.1-8B, and Qwen3-4B. We report WikiText-2 perplexity and MCQA task accuracy, where our \BOS-based strategy consistently outperforms Mag-SP and Wanda-SP while remaining close to dense baselines.}
\label{tab:component_pruning_results}
\end{table*}

\begin{table*}[t]
\centering
\small
\resizebox{\linewidth}{!}{
\begin{tabular}{c | c | c | c | cccccccc | c}
\toprule
\textbf{Model} & \textbf{Strategy} & \makecell[c]{\textbf{Pruning}\\\textbf{Ratio}} & \textbf{Wikitext-2 $\downarrow$} & \textbf{ARC-c} & \textbf{ARC-e} & \textbf{BoolQ} & \textbf{HellaS.} & \textbf{MMLU} & \textbf{OBQA} & \textbf{PIQA} & \textbf{WG} & \textbf{Avg.} \\
\midrule\midrule
\multirow{7}{*}{Gemma-3-4B} & Dense & 0 & 10.21 & 0.533 & 0.816 & 0.791 & 0.580 & 0.597 & 0.352 & 0.788 & 0.725 & 0.648 \\
\cmidrule{2-13}
 & Bottom Up & \multirow{6}{*}{0.125} & 50643.64 & 0.206 & 0.314 & 0.466 & 0.260 & 0.241 & 0.146 & 0.549 & 0.488 & 0.334 \\
 & Top Down &  & 45.68 & 0.379 & 0.629 & 0.254 & 0.427 & 0.178 & 0.274 & 0.717 & \underline{0.657} & 0.439 \\
 & ShortGPT &  & 20.82 & 0.307 & 0.673 & 0.412 & 0.431 & 0.283 & 0.254 & 0.743 & 0.549 & 0.456 \\
 & Mag-Layer &  & 33.28 & 0.213 & 0.627 & 0.619 & 0.349 & 0.269 & 0.242 & 0.705 & 0.531 & 0.444 \\
 & Wanda-Layer &  & $\mathbf{15.18}$ & \underline{0.400} & $\mathbf{0.739}$ & \underline{0.633} & \underline{0.478} & \underline{0.300} & \underline{0.280} & $\mathbf{0.762}$ & 0.564 & \underline{0.519} \\
 &\textbf{Ours (\texttt{<BOS>})} &  & \underline{20.02} & $\mathbf{0.456}$ & \underline{0.715} & $\mathbf{0.701}$ & $\mathbf{0.507}$ & $\mathbf{0.584}$ & $\mathbf{0.300}$ & \underline{0.749} & $\mathbf{0.701}$ & $\mathbf{0.589}$ \\
\midrule
\multirow{7}{*}{Gemma-3-12B} & Dense & 0 & 10.18 & 0.650 & 0.872 & 0.854 & 0.645 & 0.741 & 0.354 & 0.807 & 0.803 & 0.716 \\
\cmidrule{2-13}
 & Bottom Up & \multirow{6}{*}{0.250} & 3968.23 & 0.189 & 0.312 & 0.497 & 0.262 & 0.250 & 0.156 & 0.542 & 0.509 & 0.340 \\
 & Top Down &  & \underline{191.38} & \underline{0.407} & \underline{0.628} & 0.256 & \underline{0.365} & \underline{0.452} & \underline{0.272} & \underline{0.670} & \underline{0.681} & \underline{0.466} \\
 & ShortGPT &  & 305692485.89 & 0.228 & 0.342 & 0.433 & 0.268 & 0.234 & 0.172 & 0.563 & 0.489 & 0.341 \\
 & Mag-Layer &  & 173294.34 & 0.261 & 0.452 & \underline{0.506} & 0.310 & 0.249 & 0.188 & 0.641 & 0.511 & 0.390 \\
 & Wanda-Layer &  & 18253938.76 & 0.245 & 0.350 & 0.454 & 0.295 & 0.236 & 0.190 & 0.553 & 0.481 & 0.351 \\
 &\textbf{Ours (\texttt{<BOS>})} &  & $\mathbf{63.98}$ & $\mathbf{0.430}$ & $\mathbf{0.665}$ & $\mathbf{0.655}$ & $\mathbf{0.450}$ & $\mathbf{0.501}$ & $\mathbf{0.274}$ & $\mathbf{0.708}$ & $\mathbf{0.690}$ & $\mathbf{0.547}$ \\
\midrule
\multirow{7}{*}{Llama-3.1-8B} & Dense & 0 & 7.71 & 0.550 & 0.822 & 0.829 & 0.618 & 0.653 & 0.336 & 0.791 & 0.781 & 0.673 \\
\cmidrule{2-13}
 & Bottom Up & \multirow{6}{*}{0.250} & 129826.91 & 0.206 & 0.270 & 0.456 & 0.260 & 0.228 & 0.132 & 0.536 & 0.493 & 0.323 \\
 & Top Down &  & $\mathbf{1202.52}$ & $\mathbf{0.270}$ & 0.362 & $\mathbf{0.622}$ & 0.249 & $\mathbf{0.523}$ & $\mathbf{0.220}$ & 0.578 & $\mathbf{0.582}$ & $\mathbf{0.426}$ \\
 & ShortGPT &  & 7211.05 & \underline{0.269} & \underline{0.436} & 0.375 & \underline{0.271} & \underline{0.342} & 0.194 & \underline{0.614} & \underline{0.542} & 0.380 \\
 & Mag-Layer &  & 136054.85 & 0.201 & 0.264 & 0.383 & 0.260 & 0.243 & 0.116 & 0.540 & 0.501 & 0.313 \\
 & Wanda-Layer &  & 85784.66 & 0.210 & 0.276 & 0.402 & 0.259 & 0.230 & 0.132 & 0.532 & 0.477 & 0.315 \\
 &\textbf{Ours (\texttt{<BOS>})} &  & \underline{2017.63} & 0.212 & $\mathbf{0.518}$ & \underline{0.478} & $\mathbf{0.284}$ & 0.266 & \underline{0.210} & $\mathbf{0.635}$ & 0.521 & \underline{0.390} \\
\midrule
\multirow{7}{*}{Qwen3-4B} & Dense & 0 & 17.31 & 0.584 & 0.804 & 0.851 & 0.532 & 0.701 & 0.294 & 0.753 & 0.672 & 0.649 \\
\cmidrule{2-13}
 & Bottom Up & \multirow{6}{*}{0.125} & 120.72 & 0.260 & 0.513 & 0.572 & 0.301 & 0.293 & 0.178 & 0.586 & 0.499 & 0.400 \\
 & Top Down &  & 72.55 & 0.408 & 0.614 & \underline{0.717} & 0.392 & $\mathbf{0.682}$ & 0.236 & 0.648 & \underline{0.633} & 0.541 \\
 & ShortGPT &  & \underline{41.62} & $\mathbf{0.460}$ & 0.662 & 0.564 & $\mathbf{0.450}$ & \underline{0.673} & \underline{0.240} & $\mathbf{0.680}$ & 0.623 & \underline{0.544} \\
 & Mag-Layer &  & $\mathbf{39.67}$ & 0.428 & $\mathbf{0.670}$ & 0.708 & 0.403 & 0.475 & 0.218 & \underline{0.677} & 0.572 & 0.519 \\
 & Wanda-Layer &  & 6686.82 & 0.195 & 0.275 & 0.388 & 0.260 & 0.253 & 0.124 & 0.540 & 0.458 & 0.312 \\
 &\textbf{Ours (\texttt{<BOS>})} &  & 49.89 & \underline{0.451} & \underline{0.666} & $\mathbf{0.826}$ & \underline{0.441} & 0.611 & $\mathbf{0.252}$ & 0.674 & $\mathbf{0.642}$ & $\mathbf{0.570}$ \\
\bottomrule
\end{tabular}
}
\caption{Performance comparison of layer-level pruning strategies on Gemma-3-4B/12B, Llama-3.1-8B, and Qwen3-4B. Mag-Layer and Wanda-Layer are block-level adaptations of Mag-SP and Wanda-SP. Bold and underline denote the best and second-best result per column within each model. Our \BOS-based strategy attains the best or second-best MCQA average in every setting, whereas the weight- and activation-derived baselines are competitive on perplexity only.}
\label{tab:layer_pruning_results}
\end{table*}

\section{Experiments}

\subsection{Experimental Setup}

\paragraph{Models and Benchmarks}
We perform evaluations using three LLMs that employ Grouped-Query Attention (GQA): Gemma-3-4B/12B \citep{gemmateam2025gemma3technicalreport}, Llama-3.1-8B \citep{grattafiori2024llama3herdmodels}, and Qwen3-4B \citep{yang2025qwen3technicalreport}. All experiments are conducted with a maximum sequence length of 4,096 tokens.

To assess performance, we use the lm-evaluation-harness\footnote{https://github.com/EleutherAI/lm-evaluation-harness} \citep{lintang_sutawika_2023_10256836} across multiple benchmark groups. For core language understanding, we evaluate WikiText-2 \citep{merity2017pointer}, ARC-Easy \citep{Clark2018ThinkYH}, BoolQ \citep{clark-etal-2019-boolq}, OpenbookQA \citep{mihaylov-etal-2018-suit}, and PIQA \citep{DBLP:conf/aaai/BiskZLGC20}. For reasoning-focused evaluation, we use Winogrande \citep{sakaguchi2019winogrande}, MMLU \citep{hendrycks2021measuring}, HellaSwag \citep{zellers2019hellaswag}, and ARC-Challenge \citep{Clark2018ThinkYH}. For generation, we evaluate HumanEval \citep{chen2021evaluating}, TriviaQA \citep{joshi2017triviaqa}, and GSM8K \citep{cobbe2021training}. For long-context and retrieval, we use LongBench \citep{bai2023longbench} and report HotpotQA \citep{yang2018hotpotqa}, QASPER \citep{dasigi2021qasper}, and GovReport \citep{huang2021efficient}, following the LongBench configuration. Detailed dataset settings are provided in Appendix~\ref{sec:dataset_summary}.

\paragraph{Baselines and Implementation}
We compare our method with Wanda-SP \citep{an2023fluctuation}, Mag-SP \citep{an2023fluctuation}, and ShortGPT \citep{men2025shortgpt}. For layer-level pruning, we also include two positional baselines: Bottom-Up (pruning from early to late layers) and Top-Down (from late to early layers), to test whether redundancy reflects functional importance or layer position. In all settings, we keep the first and last layers to preserve stable input feature extraction and output representation.

Because the FLAP implementations of Wanda-SP and Mag-SP\footnote{https://github.com/CASIA-LMC-Lab/FLAP} do not natively support Grouped-Query Attention (GQA), we adapted their core logic for GQA and reimplemented all baselines under the same framework.

Because Wanda-SP and Mag-SP are component-level criteria, they do not define an ordering over full transformer blocks in their original formulation. To obtain layer-level counterparts, we aggregate their per-component importance scores into a single scalar per block by averaging over all output channels of the attention and MLP projections in that block, and then remove the blocks with the lowest aggregated importance. We denote these adapted baselines Wanda-Layer and Mag-Layer. Full implementation details of Wanda-SP and Mag-SP, together with a discussion of how the structured variants relate to their unstructured counterparts, are given in Appendix~\ref{sec:wandasp_details}.

\begin{table}[!t]
\centering
\small
\resizebox{\linewidth}{!}{
\begin{tabular}{c | c | c | ccc}
\toprule
\textbf{Model} & \textbf{Strategy} & \makecell[c]{\textbf{Pruning}\\\textbf{Ratio}} & \textbf{GSM8K} & \textbf{HumanEval} & \textbf{TriviaQA} \\
\midrule\midrule
\multirow{4}{*}{Gemma-3-4B} & Dense & 0 & 0.392 & 0.348 & 0.605 \\
\cmidrule{2-6}
 & Mag-sp & \multirow{3}{*}{0.125} & 0.010 & 0.049 & 0.489 \\
 & Wanda-sp &  & 0.024 & 0.171 & 0.571 \\
 & \textbf{Ours (\texttt{<BOS>})} &  & $\mathbf{0.263}$ & $\mathbf{0.341}$ & $\mathbf{0.588}$ \\
\midrule
\multirow{4}{*}{Gemma-3-12B} & Dense & 0 & 0.715 & 0.427 & 0.747 \\
\cmidrule{2-6}
 & Mag-sp & \multirow{3}{*}{0.125} & 0.040 & 0.189 & 0.674 \\
 & Wanda-sp &  & 0.287 & 0.250 & 0.682 \\
 & \textbf{Ours (\texttt{<BOS>})} &  & $\mathbf{0.667}$ & $\mathbf{0.402}$ & $\mathbf{0.714}$ \\
\midrule
\multirow{4}{*}{Llama-3.1-8B} & Dense & 0 & 0.491 & 0.366 & 0.704 \\
\cmidrule{2-6}
 & Mag-sp & \multirow{3}{*}{0.125} & 0 & 0.073 & 0.517 \\
 & Wanda-sp &  & 0.150 & 0.274 & $\mathbf{0.687}$ \\
 & \textbf{Ours (\texttt{<BOS>})} &  & $\mathbf{0.445}$ & $\mathbf{0.329}$ & 0.634 \\
\midrule
\multirow{4}{*}{Qwen3-4B} & Dense & 0 & 0.861 & 0.750 & 0.442 \\
\cmidrule{2-6}
 & Mag-sp & \multirow{3}{*}{0.125} & 0.017 & 0.134 & 0.232 \\
 & Wanda-sp &  & 0.188 & 0.494 & $\mathbf{0.408}$ \\
 & \textbf{Ours (\texttt{<BOS>})} &  & $\mathbf{0.317}$ & $\mathbf{0.543}$ & 0.376 \\
\bottomrule
\end{tabular}
}
\caption{Performance comparison of generation tasks under component-level pruning on Gemma-3-4B/12B, Llama-3.1-8B, and Qwen3-4B. We report GSM8K and TriviaQA accuracy, and HumanEval Pass@1.}
\label{tab:generation_component_results}
\end{table}

\subsection{Perplexity and MCQA Results
}

\paragraph{Component-level Pruning}
Table \ref{tab:component_pruning_results} summarizes component-level pruning results. Our \texttt{<BOS>}-based method achieves the highest MCQA average among the pruning baselines while remaining close to the dense baselines.

At 12.5\% pruning, \texttt{<BOS>}-based pruning achieves average accuracies of 0.641 and 0.643 on Gemma-3-4B and Qwen3-4B, respectively, compared with dense baselines of 0.648 and 0.649. At 25.0\%, it also performs best on Gemma-3-12B (0.694) and Llama-3.1-8B (0.631), consistently outperforming Mag-SP and Wanda-SP. Although Wanda-SP occasionally yields lower WikiText-2 perplexity, our method better preserves MCQA accuracy, suggesting that \BOS sink scores are a stronger criterion for maintaining reasoning performance under structured pruning. Additional results are provided in Appendix~\ref{sec:robust_pruning_anaylsis}.

\paragraph{Perplexity and Task Retention}
Perplexity degrades more sharply than task accuracy under component-level pruning. Under our
criterion it stays within one to three points of the dense baseline on Gemma-3-4B, Gemma-3-12B, and
Qwen3-4B, whereas Llama-3.1-8B rises from 7.71 to 20.55 at 25\%. The two measures also order the
criteria differently: on Llama-3.1-8B, Wanda-SP attains a lower perplexity than our method while
retaining less MCQA accuracy, and at 12.5\% it also trails on GSM8K and HumanEval, leading only on
TriviaQA. Perplexity on a single held-out corpus is therefore a limited proxy for task-relevant
capability under structured pruning, which is why we evaluate generation
(Table~\ref{tab:generation_component_results}) and long-context
(Table~\ref{tab:longbench_component_results}) performance alongside it. The magnitude of the
perplexity increase and its dependence on pruning ratio and model scale are examined in
Appendix~\ref{sec:ppl_scale_ratio}.

\paragraph{Layer-level Pruning}
Table~\ref{tab:layer_pruning_results} reports layer-level pruning, where each step removes full
transformer blocks (attention + FFN). Besides the positional baselines and ShortGPT, we compare
against Mag-Layer and Wanda-Layer, the block-level adaptations. The \texttt{<BOS>}-based strategy attains the best MCQA average on
Gemma-3-4B (0.589), Gemma-3-12B (0.547), and Qwen3-4B (0.570); on Llama-3.1-8B it surpasses every
baseline except Top-Down (0.390 vs.\ 0.426).

The two adapted baselines separate language-modeling loss from task-relevant capability. Wanda-Layer
attains the lowest perplexity on Gemma-3-4B (15.18 vs.\ our 20.02) and Mag-Layer on Qwen3-4B (39.67
vs.\ our 49.89), yet both trail our MCQA average by 7 and 5 points; at 25.0\% they collapse to
perplexities of $10^{4}$--$10^{7}$ and averages of 0.390/0.351 (Gemma-3-12B) and 0.313/0.315
(Llama-3.1-8B), against 0.547 and 0.390 for ours. Averaging a per-channel saliency into a block score
thus transfers poorly across depth. The layer-wise \BOS score instead measures the share of a
block's attention mass that is allocated to the anchor token, a quantity we relate to ablation
insensitivity in Section~\ref{sec:bos_redundancy}, and it is the only criterion in
Table~\ref{tab:layer_pruning_results} never worse than second on MCQA average.

\subsection{Generation Results}

Table~\ref{tab:generation_component_results} summarizes generation performance under component-level pruning. We report GSM8K and TriviaQA accuracy, and HumanEval Pass@1. Overall, \texttt{<BOS>}-based pruning is substantially more robust than Mag-SP and remains competitive with Wanda across model families.

Across all four models, \BOS yields the highest GSM8K and HumanEval scores among the pruned variants, with a particularly wide margin over Mag-SP, which reduces GSM8K accuracy to near zero in every case. TriviaQA behaves differently: \BOS leads on both Gemma models but trails Wanda-SP on Llama-3.1-8B (0.634 vs.\ 0.687) and Qwen3-4B (0.376 vs.\ 0.408), the two families for which Wanda-SP also attains the lower perplexity in Table~\ref{tab:component_pruning_results}. Retention relative to the dense model is likewise uneven, ranging from 93\% of dense GSM8K accuracy on Gemma-3-12B to 37\% on Qwen3-4B. These results indicate that \BOS sink scores preserve generation quality more reliably than the baselines: the margin over the strongest baseline exceeds 12 points on GSM8K for every model, while the two TriviaQA deficits stay within 6 points.

\subsection{Long-Context and Retrieval Results}

Table~\ref{tab:longbench_component_results} summarizes long-context and retrieval performance under component-level pruning. We report HotpotQA and QASPER F1 scores, and GovReport ROUGE. Overall, \texttt{<BOS>}-based pruning consistently outperforms Mag-SP and remains competitive with Wanda across LongBench tasks.

In particular, \texttt{<BOS>} shows strong gains on retrieval-oriented tasks for Gemma-3-4B and Llama-3.1-8B, including matching the dense QASPER score on Llama-3.1-8B at a pruning ratio of 0.125. On Qwen3-4B, our method remains robust and consistently outperforms Mag-SP, although Wanda performs better on QASPER and GovReport. This pattern indicates model-family sensitivity in the reliability of the \texttt{<BOS>} pruning signal.

\begin{table}[!t]
\centering
\small
\resizebox{\linewidth}{!}{
\begin{tabular}{c | c | c | ccc}
\toprule
\textbf{Model} & \textbf{Strategy} & \textbf{Pruning Ratio} & \textbf{HotpotQA} & \textbf{QASPER} & \textbf{GovReport} \\
\midrule\midrule
\multirow{4}{*}{Gemma-3-4B} & Dense & 0 & 0.071 & 0.101 & 0.270 \\
\cmidrule{2-6}
 & Mag-sp & \multirow{3}{*}{0.125} & 0.015 & 0.023 & 0.071 \\
 & Wanda-sp &  & 0.047 & 0.043 & $\mathbf{0.153}$ \\
 & \textbf{Ours (\texttt{<BOS>})} &  & $\mathbf{0.048}$ & $\mathbf{0.063}$ & 0.140 \\
\midrule
\multirow{4}{*}{Llama-3.1-8B} & Dense & 0 & 0.108 & 0.134 & 0.302 \\
\cmidrule{2-6}
 & Mag-sp & \multirow{3}{*}{0.125} & 0.043 & 0.048 & 0.098 \\
 & Wanda-sp &  & 0.087 & 0.090 & 0.191 \\
 & \textbf{Ours (\texttt{<BOS>})} &  & $\mathbf{0.093}$ & $\mathbf{0.134}$ & $\mathbf{0.237}$ \\
\midrule
\multirow{4}{*}{Qwen3-4B} & Dense & 0 & 0.128 & 0.104 & 0.300 \\
\cmidrule{2-6}
 & Mag-sp & \multirow{3}{*}{0.125} & 0.034 & 0.043 & 0.132 \\
 & Wanda-sp &  & 0.069 & $\mathbf{0.100}$ & $\mathbf{0.247}$ \\
 & \textbf{Ours (\texttt{<BOS>})} &  & $\mathbf{0.084}$ & 0.087 & 0.235 \\
\bottomrule
\end{tabular}
}
\caption{Performance comparison of long-context and retrieval tasks under component-level pruning on Gemma-3-4B, Llama-3.1-8B, and Qwen3-4B. We report HotpotQA and QASPER F1 scores, and GovReport ROUGE.}
\label{tab:longbench_component_results}
\end{table}

\begin{figure*}[t]
    \centering
    \includegraphics[width=0.32\linewidth]{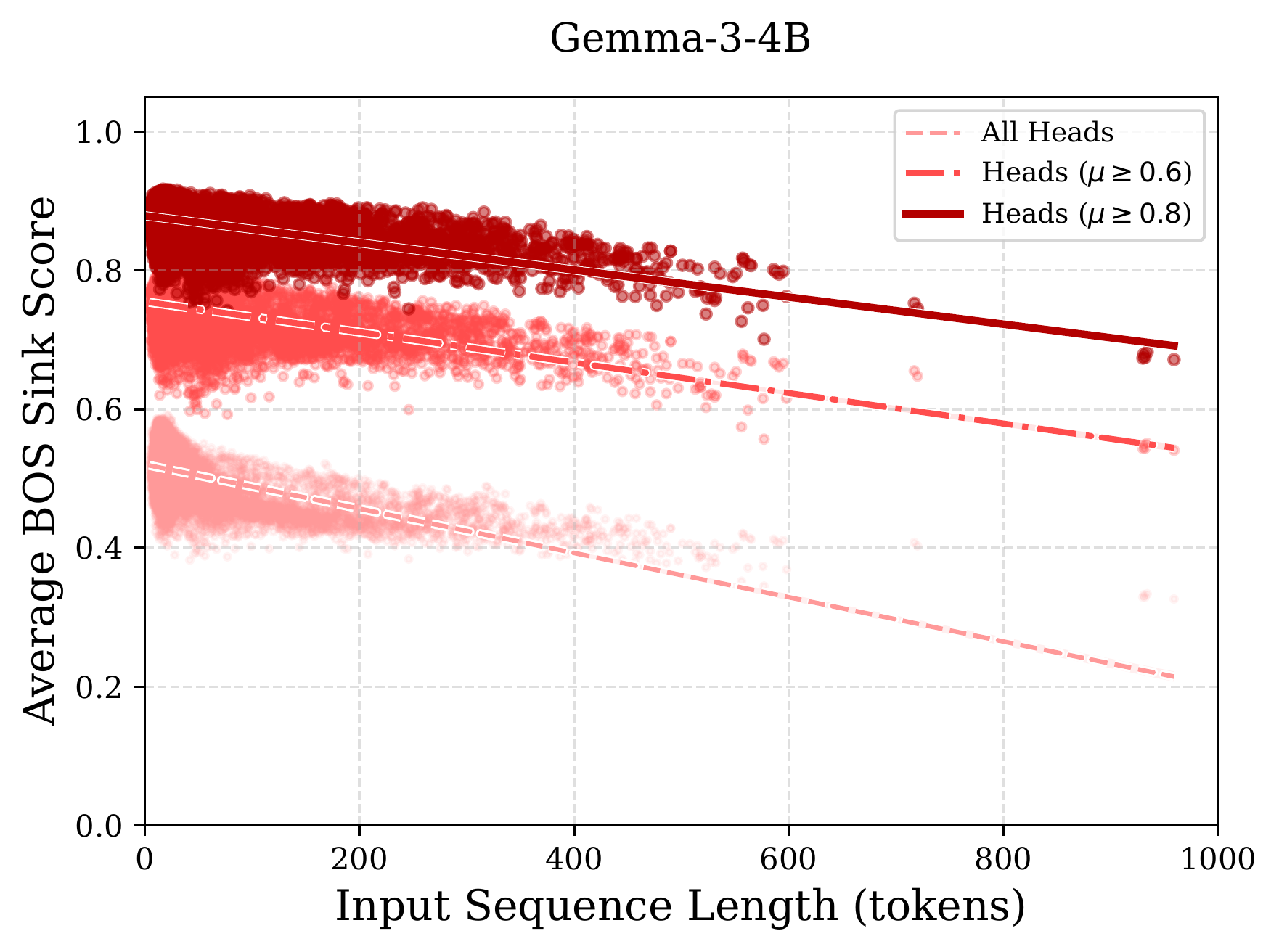}
    \includegraphics[width=0.32\linewidth]{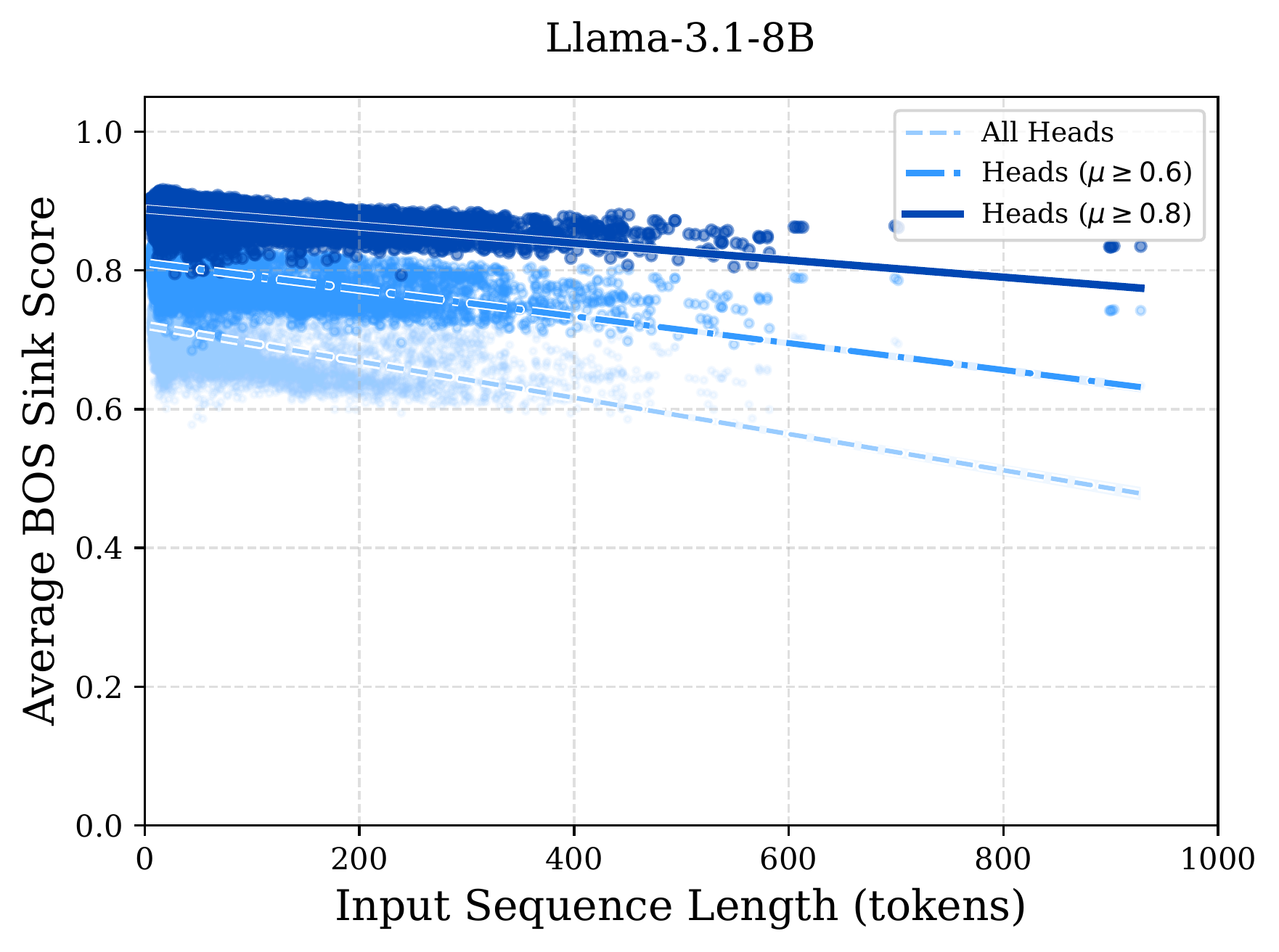}
    \includegraphics[width=0.32\linewidth]{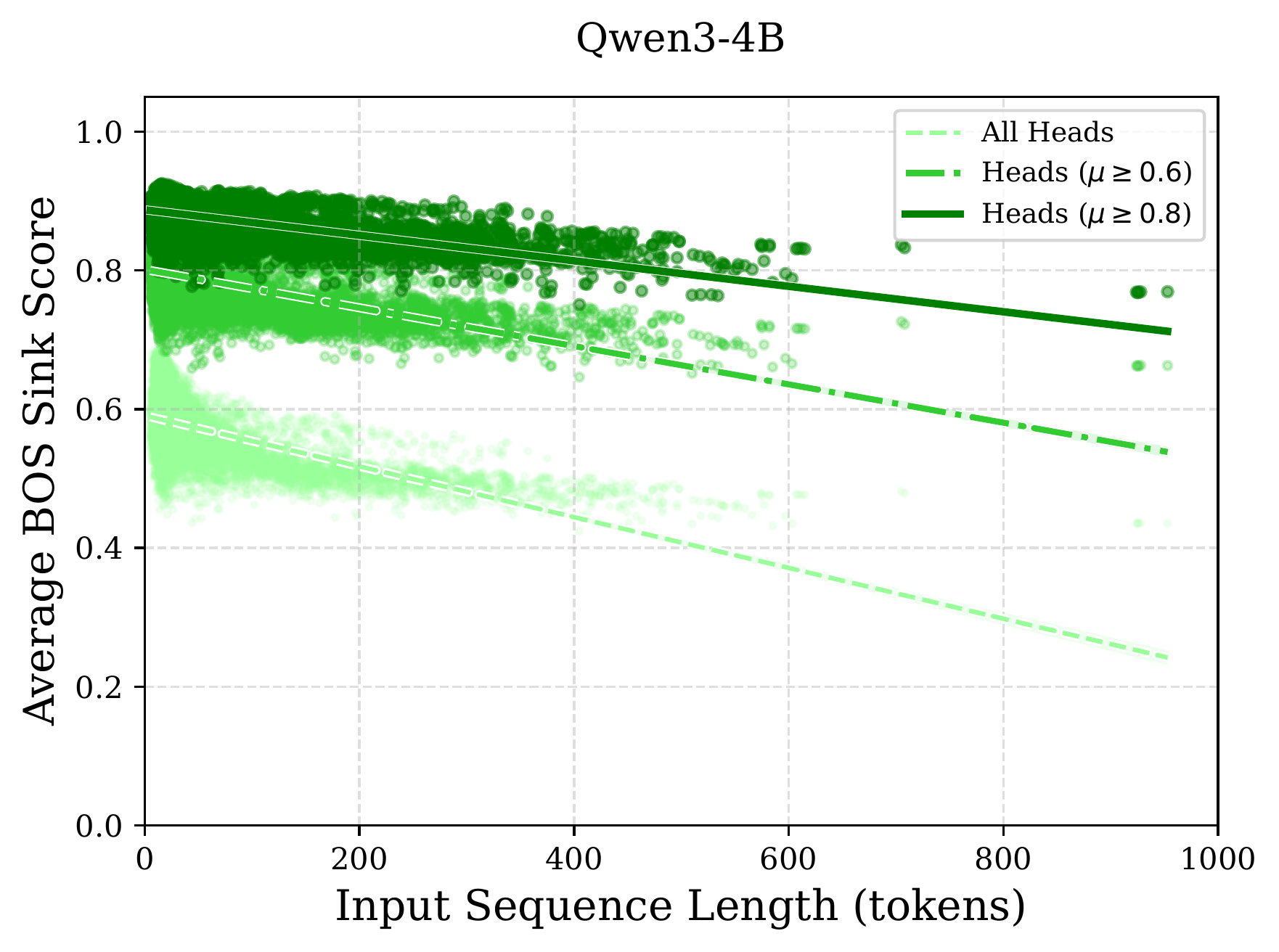}
    \caption{Mean \BOS sink score versus sequence length $T$. Points show input-level means, and dashed, dash-dotted, and solid lines are linear fits for all heads, $\mu \ge 0.6$, and $\mu \ge 0.8$, respectively. Although softmax implies $1/T$ scaling, higher-$\mu$ sink heads show shallower slopes, indicating more stable \BOS focus at longer contexts.}
    \label{fig:bos_scatter}
\end{figure*}

\begin{figure*}[t]
    \centering
    \includegraphics[width=0.32\linewidth]{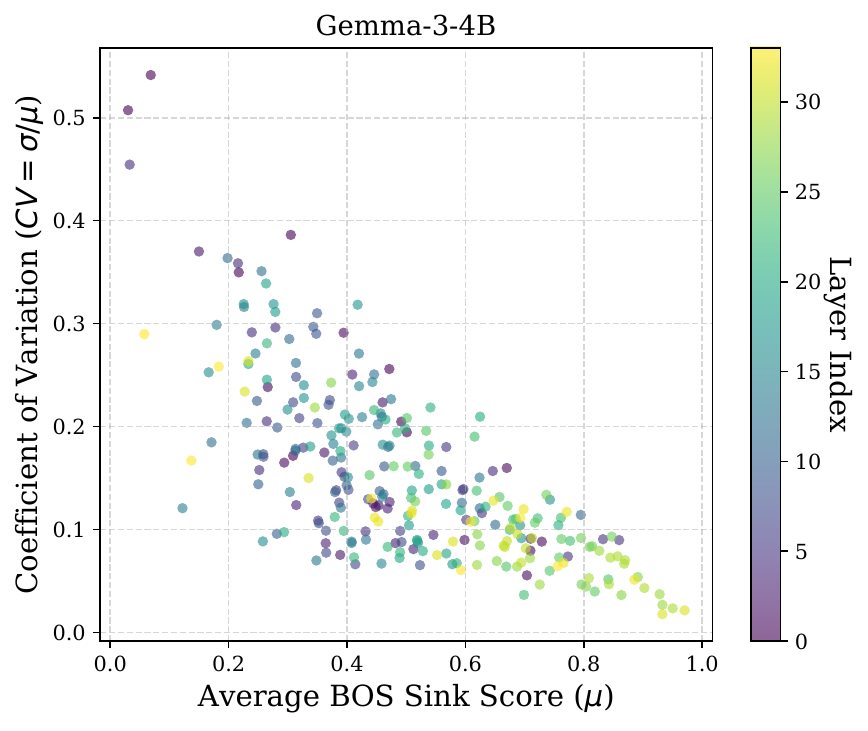}
    \includegraphics[width=0.32\linewidth]{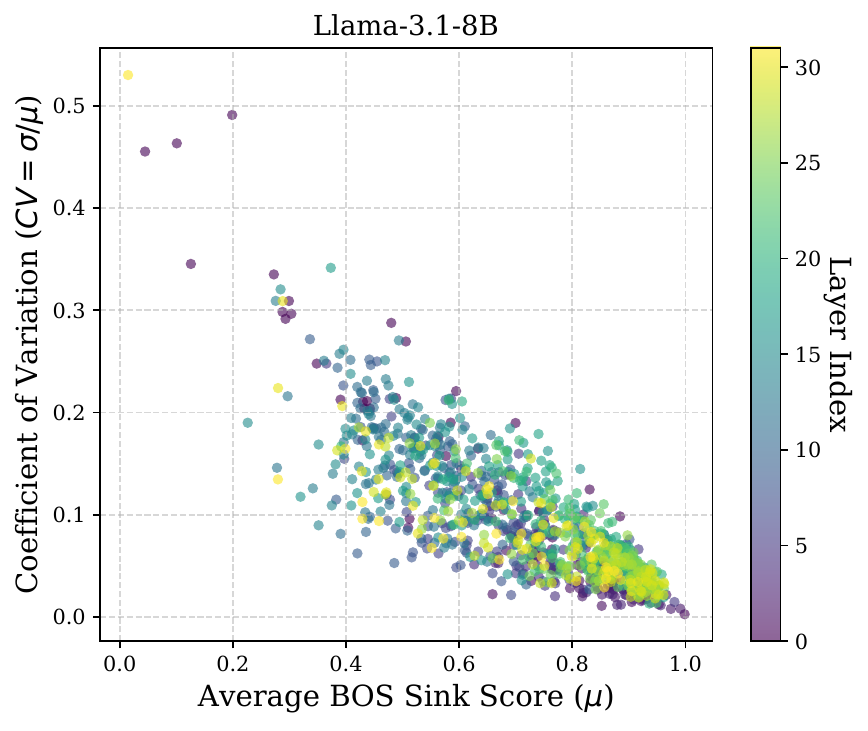}
    \includegraphics[width=0.32\linewidth]{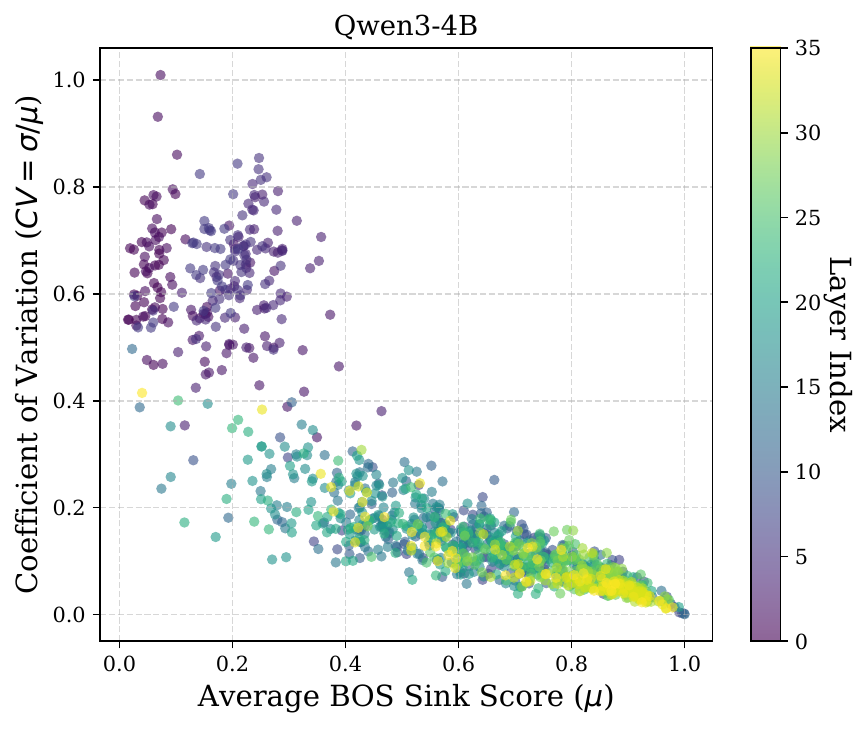}
    \caption{Relationship between mean \BOS sink score ($\mu$) and coefficient of variation ($CV$). Each dot denotes an attention head, colored by layer. Higher-$\mu$ heads exhibit lower $CV$ and are concentrated in deeper layers, indicating stable sink behavior and layer-wise redundancy.}
    \label{fig:cv_head_scatter}
\end{figure*}

\section{Validation of the \BOS Sink-based Pruning Criterion}

\subsection{Distinctiveness of \BOS-based Head Selection}

To directly examine whether our pruning criterion selects a head set that is truly distinct from existing methods, we compute the intersection-over-union (IoU) between the attention heads pruned by our method and those selected by Mag-SP and Wanda-SP under component-level pruning. As shown in Table~\ref{tab:iou_head_overlap}, our method has consistently low overlap with Mag-SP (0.125, 0.113, and 0.104 for Gemma-3-4B, Llama-3.1-8B, and Qwen3-4B, respectively). Although overlap with Wanda-SP is higher (0.219, 0.453, and 0.431), a substantial portion of heads remains non-overlapping. These results suggest that the \BOS sink signal captures structural redundancy that is not fully explained by weight magnitude or activation-based criteria alone. Additional experimental details are presented in Appendix ~\ref{sec:extended_iou_overlap}.

\begin{table}[t]
\centering
\small
\resizebox{\linewidth}{!}{
\begin{tabular}{l c c c}
\toprule
\textbf{Model} & \textbf{Pruning Ratio} & \textbf{Mag-sp} & \textbf{Wanda-sp} \\
\midrule
\midrule
Gemma-3-4B & 0.125 & 0.125 & 0.219 \\
Llama-3.1-8B & 0.25 & 0.113 & 0.453 \\
Qwen3-4B & 0.125 & 0.104 & 0.431 \\
\bottomrule
\end{tabular}
}
\caption{IoU of pruned attention heads between our method and Mag-SP/Wanda-SP under component-level pruning.}
\label{tab:iou_head_overlap}
\end{table}

\subsection{Consistency Across Context Lengths}

To further validate the reliability of \BOS sink scores as a pruning metric, we investigate their stability across varying input sequence lengths $T$.

\paragraph{Scaling and Structural Normalization}
By definition, the \BOS sink score represents the average of attention weights assigned to the \BOS token throughout a sequence. Due to the Softmax operation's normalization---where the sum of attention weights for each query must equal 1---the average attention weight per token naturally scales with $1/T$. Consequently, as $T$ increases, absolute \BOS sink scores exhibit an inevitable downward trend across all heads. This reduction should be viewed as a structural consequence of the attention mechanism's normalization rather than a loss of functional importance.

\paragraph{Differential Stability of Sink Heads}
Notably, we observe a clear correlation between a head's average \BOS sink score $\mu$ and its robustness to increasing sequence lengths. As illustrated in Figure~\ref{fig:bos_scatter}, the rate of decline in attention scores is inversely related to the magnitude of $\mu$: while the average score for all heads drops sharply, heads with $\mu \ge 0.6$ show more tempered decreases, and those with $\mu \ge 0.8$ exhibit the most gradual downward trajectories. This differential stability indicates that higher-scoring \BOS sink heads appear to partially mitigate the $1/T$ scaling effect, maintaining a disproportionately high focus on the \BOS token even as the total attention budget is distributed over an increasingly large number of tokens.

\paragraph{Positional Bias and Convergence}
Furthermore, our analysis reveals a strong positional bias in the emergence and stability of these heads. To quantify this, we use the coefficient of variation ($CV = \sigma/\mu$), where $\sigma$ and $\mu$ represent the standard deviation and mean of the \BOS sink scores across different sequence lengths. As shown in Figure~\ref{fig:cv_head_scatter}, a clear inverse correlation exists between $\mu$ and $CV$. Crucially, heads with high $\mu$ and near-zero $CV$ are predominantly concentrated in the deeper layers of the architecture (indicated by the yellow and green hues). This suggests that as the model's layers deepen, certain heads converge into stable \BOS sinks that maintain consistent attention patterns regardless of the input sequence length.

\section{Conclusion}
In this study, we established the \BOS sink phenomenon as a primary driver of functional redundancy in LLMs, revealing a strong correlation between high \BOS sink scores and negligible predictive contribution. Our findings provide a systematic explanation for the high redundancy observed in deeper layers, where specialized heads act as stable repositories for unnecessary attention weights, effectively functioning as internal \emph{dumping grounds}. By bridging the gap between internal attention patterns and layer-wise sensitivity, we demonstrated that structural attention properties can serve as a more efficient and direct alternative to traditional weight-based metrics for model compression. Experiments across Gemma-3, Llama-3.1, and Qwen3 confirm that our straightforward \texttt{<BOS>}-based pruning matches or exceeds the efficacy of established methods like Wanda-SP and Mag-SP.

Future work will focus on validating the \BOS sink phenomenon across different attention
architectures, including MoE and MHA, as well as under modified attention operators that give heads
an explicit outlet for surplus attention mass, to assess whether sink behavior is
architecture-dependent. Furthermore, we aim to investigate whether similar attention-anchoring mechanisms exist in other Transformer-based architectures beyond text, such as Vision-Language Models (VLMs). Additionally, investigating the developmental dynamics of these sinks during the pre-training phase may offer deeper insights into how models establish representational stability through structural redundancy.

\newpage
\section*{Limitations}

Despite the robust findings presented in this study, several limitations remain.

First, while we established a strong correlation between high \BOS sink scores and functional redundancy, our work is primarily based on empirical observations. We demonstrated that these specialized heads predominantly emerge and concentrate in the deeper layers of the architecture, yet we do not provide a formal theoretical derivation for the underlying cause of this positional concentration.

Second, although we validated the efficacy of our \BOS-based pruning strategy across diverse GQA-based architectures---specifically Gemma-3-4B/12B, Llama-3.1-8B, and Qwen3-4B---further experiments are needed on ultra-large-scale models (e.g., 70B and above) and on a broader range of attention architectures beyond GQA, including MoE and MHA, to verify whether these attention patterns and redundancy characteristics generalize.

Third, structured pruning removes entire architectural units rather than individual weights, which is what makes the efficiency gains of Appendix~\ref{sec:efficiency_analysis_layer_pruning} realizable but also makes the intervention coarse. Degradation therefore grows quickly with the pruning ratio, most visibly in perplexity: on Llama-3.1-8B at 25\%, every layer-level strategy including ours exceeds a perplexity of 1{,}000 while MCQA accuracy remains relatively stable. Our results support the \BOS sink score in the low-to-moderate regime rather than under aggressive compression.

Fourth, the \BOS sink score presupposes the standard softmax, whose normalization constraint forces every query's weights to sum to one and therefore requires surplus mass to land on an existing token. In architectures that supply an explicit outlet instead---for instance, a constant in the softmax denominator \citep{miller2023attention}---the surplus need not accumulate at position 0, and our score would return uniformly low values and lose its ability to rank heads. Whether the redundancy it tracks disappears in such models, or merely moves to that outlet, remains untested.


\section*{Ethical Considerations}
\paragraph{Fairness and Potential Bias}
The pruning process, by removing model parameters, carries the risk of inadvertently amplifying biases toward specific demographic groups inherent in the training data.
While our study demonstrates robust performance retention on general reasoning benchmarks such as MMLU and ARC, the specific impact on social bias and toxicity remains to be fully characterized.
Consequently, we strongly advise conducting a comprehensive evaluation to detect any discriminatory behaviors against marginalized communities before deploying the pruned models in production environments.

\paragraph{Use of AI Assistants}
In accordance with the ACL Policy on AI Assistance, we acknowledge the use of Gemini 3 Pro\footnote{https://deepmind.google/models/gemini/} to assist with code debugging and improving the readability and grammar of this manuscript.
The final content and responsibility for the work rest solely with the authors.

\section*{Acknowledgements}
This work was supported by the National Research Foundation of Korea (NRF)
grant funded by the Korea government (MSIT) (No. RS-2024-00345809, Research
on AI Robustness Against Distribution Shift in Real-World Scenarios; and
No. RS-2023-00222663, Center for Optimizing Hyperscale AI Models and
Platforms), and the Institute of Information \& Communications Technology
Planning \& Evaluation (IITP) grant funded by the Korea government (MSIT)
(No. RS-2026-25507282, Development of AI learning technologies based on
brain cognitive science and biological learning principles).

\bibliography{custom}

@inproceedings{vaswani2017attention,
 author = {Vaswani, Ashish and Shazeer, Noam and Parmar, Niki and Uszkoreit, Jakob and Jones, Llion and Gomez, Aidan N and Kaiser, \L ukasz and Polosukhin, Illia},
 booktitle = {Advances in Neural Information Processing Systems},
 editor = {I. Guyon and U. Von Luxburg and S. Bengio and H. Wallach and R. Fergus and S. Vishwanathan and R. Garnett},
 pages = {},
 publisher = {Curran Associates, Inc.},
 title = {Attention is All you Need},
 url = {https://proceedings.neurips.cc/paper_files/paper/2017/file/3f5ee243547dee91fbd053c1c4a845aa-Paper.pdf},
 volume = {30},
 year = {2017}
}

@inproceedings{voita-etal-2019-analyzing,
    title = "Analyzing Multi-Head Self-Attention: Specialized Heads Do the Heavy Lifting, the Rest Can Be Pruned",
    author = "Voita, Elena  and
      Talbot, David  and
      Moiseev, Fedor  and
      Sennrich, Rico  and
      Titov, Ivan",
    editor = "Korhonen, Anna  and
      Traum, David  and
      M{\`a}rquez, Llu{\'i}s",
    booktitle = "Proceedings of the 57th Annual Meeting of the Association for Computational Linguistics",
    month = jul,
    year = "2019",
    address = "Florence, Italy",
    publisher = "Association for Computational Linguistics",
    url = "https://aclanthology.org/P19-1580/",
    doi = "10.18653/v1/P19-1580",
    pages = "5797--5808"
}

@article{hoefler2021sparsity,
  author = {Hoefler, Torsten and Alistarh, Dan and Ben-Nun, Tal and Dryden, Nikoli and Peste, Alexandra},
    title = {Sparsity in deep learning: pruning and growth for efficient inference and training in neural networks},
    year = {2021},
    issue_date = {January 2021},
    publisher = {JMLR.org},
    volume = {22},
    number = {1},
    issn = {1532-4435},
    journal = {J. Mach. Learn. Res.},
    month = jan,
    articleno = {241},
    numpages = {124}
}

@inproceedings{brown2020language,
 author = {Brown, Tom and Mann, Benjamin and Ryder, Nick and Subbiah, Melanie and Kaplan, Jared D and Dhariwal, Prafulla and Neelakantan, Arvind and Shyam, Pranav and Sastry, Girish and Askell, Amanda and Agarwal, Sandhini and Herbert-Voss, Ariel and Krueger, Gretchen and Henighan, Tom and Child, Rewon and Ramesh, Aditya and Ziegler, Daniel and Wu, Jeffrey and Winter, Clemens and Hesse, Chris and Chen, Mark and Sigler, Eric and Litwin, Mateusz and Gray, Scott and Chess, Benjamin and Clark, Jack and Berner, Christopher and McCandlish, Sam and Radford, Alec and Sutskever, Ilya and Amodei, Dario},
 booktitle = {Advances in Neural Information Processing Systems},
 editor = {H. Larochelle and M. Ranzato and R. Hadsell and M.F. Balcan and H. Lin},
 pages = {1877--1901},
 publisher = {Curran Associates, Inc.},
 title = {Language Models are Few-Shot Learners},
 url = {https://proceedings.neurips.cc/paper_files/paper/2020/file/1457c0d6bfcb4967418bfb8ac142f64a-Paper.pdf},
 volume = {33},
 year = {2020}
}

@article{zhao2023survey,
    title={A Survey of Large Language Models},
    volume={20},
    ISSN={2095-2236},
    url={http://dx.doi.org/10.1007/s11704-026-60308-3},
    DOI={10.1007/s11704-026-60308-3},
    number={12},
    journal={Frontiers of Computer Science},
    publisher={Springer Science and Business Media LLC},
    author={Zhao, Wayne Xin and Zhou, Kun and Li, Junyi and Tang, Tianyi and Dong, Zican and Hou, Yupeng and Zhang, Beichen and Min, Yingqian and Zhang, Junjie and Liu, Peiyu and Wang, Xiaolei and Du, Yifan and Yang, Chen and Chen, Yushuo and Chen, Zhipeng and Jiang, Jinhao and Ren, Ruiyang and Li, Yifan and Tang, Xinyu and Liu, Zikang and Hu, Yiwen and Nie, Jian-Yun and Wen, Ji-Rong},
    year={2026},
    month=May
}

@article{10.1162/tacl_a_00704,
    title = "A Survey on Model Compression for Large Language Models",
    author = "Zhu, Xunyu  and
      Li, Jian  and
      Liu, Yong  and
      Ma, Can  and
      Wang, Weiping",
    journal = "Transactions of the Association for Computational Linguistics",
    volume = "12",
    year = "2024",
    address = "Cambridge, MA",
    publisher = "MIT Press",
    url = "https://aclanthology.org/2024.tacl-1.85/",
    doi = "10.1162/tacl_a_00704",
    pages = "1556--1577"
}

@inproceedings{michel2019sixteen,
    author = {Michel, Paul and Levy, Omer and Neubig, Graham},
    booktitle = {Advances in Neural Information Processing Systems},
    editor = {H. Wallach and H. Larochelle and A. Beygelzimer and F. d\textquotesingle Alch\'{e}-Buc and E. Fox and R. Garnett},
    pages = {},
    publisher = {Curran Associates, Inc.},
    title = {Are Sixteen Heads Really Better than One?},
    url = {https://proceedings.neurips.cc/paper_files/paper/2019/file/2c601ad9d2ff9bc8b282670cdd54f69f-Paper.pdf},
    volume = {32},
    year = {2019}
}

@inproceedings{clark2019does,
    title = "What Does {BERT} Look at? An Analysis of {BERT}{'}s Attention",
    author = "Clark, Kevin  and
      Khandelwal, Urvashi  and
      Levy, Omer  and
      Manning, Christopher D.",
    editor = "Linzen, Tal  and
      Chrupa{\l}a, Grzegorz  and
      Belinkov, Yonatan  and
      Hupkes, Dieuwke",
    booktitle = "Proceedings of the 2019 ACL Workshop BlackboxNLP: Analyzing and Interpreting Neural Networks for NLP",
    month = aug,
    year = "2019",
    address = "Florence, Italy",
    publisher = "Association for Computational Linguistics",
    url = "https://aclanthology.org/W19-4828/",
    doi = "10.18653/v1/W19-4828",
    pages = "276--286"
}

@inproceedings{xiao2023efficient,
    author = {Xiao, Guangxuan and Tian, Yuandong and Chen, Beidi and Han, Song and Lewis, Mike },
    booktitle = {International Conference on Learning Representations},
    editor = {B. Kim and Y. Yue and S. Chaudhuri and K. Fragkiadaki and M. Khan and Y. Sun},
    pages = {21875--21895},
    title = {Efficient Streaming Language Models with Attention Sinks},
    url = {https://proceedings.iclr.cc/paper_files/paper/2024/file/5e5fd18f863cbe6d8ae392a93fd271c9-Paper-Conference.pdf},
    volume = {2024},
    year = {2024}
}

@misc{queipo2025attention,
    title={Attention Sinks and Compression Valleys in LLMs are Two Sides of the Same Coin}, 
    author={Enrique Queipo-de-Llano and Álvaro Arroyo and Federico Barbero and Xiaowen Dong and Michael Bronstein and Yann LeCun and Ravid Shwartz-Ziv},
    year={2026},
    eprint={2510.06477},
    archivePrefix={arXiv},
    primaryClass={cs.LG},
    url={https://arxiv.org/abs/2510.06477}, 
}

@misc{
barbero2025llms,
title={Why do LLMs attend to the first token?}, 
      author={Federico Barbero and Álvaro Arroyo and Xiangming Gu and Christos Perivolaropoulos and Michael Bronstein and Petar Veličković and Razvan Pascanu},
      year={2025},
      eprint={2504.02732},
      archivePrefix={arXiv},
      primaryClass={cs.CL},
      url={https://arxiv.org/abs/2504.02732}, 
}

@inproceedings{fan2024not,
    author = {Fan, Siqi and Jiang, Xin and Li, Xiang and Meng, Xuying and Han, Peng and Shang, Shuo and Sun, Aixin and Wang, Yequan},
    title = {Not all layers of LLMs are necessary during inference},
    year = {2025},
    isbn = {978-1-956792-06-5},
    url = {https://doi.org/10.24963/ijcai.2025/566},
    doi = {10.24963/ijcai.2025/566},
    booktitle = {Proceedings of the Thirty-Fourth International Joint Conference on Artificial Intelligence},
    articleno = {566},
    numpages = {9},
    location = {Montreal, Canada},
    series = {IJCAI '25}
}

@inproceedings{gromov2024unreasonable,
  author       = {Andrey Gromov and
                  Kushal Tirumala and
                  Hassan Shapourian and
                  Paolo Glorioso and
                  Daniel A. Roberts},
  title        = {The Unreasonable Ineffectiveness of the Deeper Layers},
  booktitle    = {The Thirteenth International Conference on Learning Representations,
                  {ICLR} 2025, Singapore, April 24-28, 2025},
  publisher    = {OpenReview.net},
  year         = {2025},
  url          = {https://openreview.net/forum?id=ngmEcEer8a},
  bibsource    = {dblp computer science bibliography, https://dblp.org}
}

@inproceedings{men2025shortgpt,
  title = "{S}hort{GPT}: Layers in Large Language Models are More Redundant Than You Expect",
    author = "Men, Xin  and
      Xu, Mingyu  and
      Zhang, Qingyu  and
      Yuan, Qianhao  and
      Wang, Bingning  and
      Lin, Hongyu  and
      Lu, Yaojie  and
      Han, Xianpei  and
      Chen, Weipeng",
    editor = "Che, Wanxiang  and
      Nabende, Joyce  and
      Shutova, Ekaterina  and
      Pilehvar, Mohammad Taher",
    booktitle = "Findings of the Association for Computational Linguistics: ACL 2025",
    month = jul,
    year = "2025",
    address = "Vienna, Austria",
    publisher = "Association for Computational Linguistics",
    url = "https://aclanthology.org/2025.findings-acl.1035/",
    doi = "10.18653/v1/2025.findings-acl.1035",
    pages = "20192--20204",
    ISBN = "979-8-89176-256-5"
}

@misc{liu2025high,
  title={High-Layer Attention Pruning with Rescaling}, 
  author={Songtao Liu and Peng Liu},
  year={2026},
  eprint={2507.01900},
  archivePrefix={arXiv},
  primaryClass={cs.CL},
  url={https://arxiv.org/abs/2507.01900}, 
}

@misc{zhang2024finercut,
    title={FinerCut: Finer-grained Interpretable Layer Pruning for Large Language Models}, 
    author={Yang Zhang and Yawei Li and Xinpeng Wang and Qianli Shen and Barbara Plank and Bernd Bischl and Mina Rezaei and Kenji Kawaguchi},
    year={2024},
    eprint={2405.18218},
    archivePrefix={arXiv},
    primaryClass={cs.LG},
    url={https://arxiv.org/abs/2405.18218}, 
}

@inproceedings{song2024sleb,
    author = {Song, Jiwon and Oh, Kyungseok and Kim, Taesu and Kim, Hyungjun and Kim, Yulhwa and Kim, Jae-Joon},
    title = {SLEB: streamlining LLMs through redundancy verification and elimination of transformer blocks},
    year = {2024},
    publisher = {JMLR.org},
    booktitle = {Proceedings of the 41st International Conference on Machine Learning},
    articleno = {1876},
    numpages = {20},
    location = {Vienna, Austria},
    series = {ICML'24}
}

@inproceedings{gu2024attention,
  author       = {Xiangming Gu and
                  Tianyu Pang and
                  Chao Du and
                  Qian Liu and
                  Fengzhuo Zhang and
                  Cunxiao Du and
                  Ye Wang and
                  Min Lin},
  title        = {When Attention Sink Emerges in Language Models: An Empirical View},
  booktitle    = {The Thirteenth International Conference on Learning Representations,
                  {ICLR} 2025, Singapore, April 24-28, 2025},
  publisher    = {OpenReview.net},
  year         = {2025},
  url          = {https://openreview.net/forum?id=78Nn4QJTEN},
  bibsource    = {dblp computer science bibliography, https://dblp.org}
}

@misc{he2024matters,
    title={What Matters in Transformers? Not All Attention is Needed}, 
    author={Shwai He and Guoheng Sun and Zheyu Shen and Ang Li},
    year={2024},
    eprint={2406.15786},
    archivePrefix={arXiv},
    primaryClass={cs.LG},
    url={https://arxiv.org/abs/2406.15786}, 
}

@misc{kaushik2025universal,
    title={The Universal Weight Subspace Hypothesis}, 
    author={Prakhar Kaushik and Shravan Chaudhari and Ankit Vaidya and Rama Chellappa and Alan Yuille},
    year={2025},
    eprint={2512.05117},
    archivePrefix={arXiv},
    primaryClass={cs.LG},
    url={https://arxiv.org/abs/2512.05117}, 
}

@inproceedings{han2015learning,
    author = {Han, Song and Pool, Jeff and Tran, John and Dally, William J.},
    title = {Learning both weights and connections for efficient neural networks},
    year = {2015},
    publisher = {MIT Press},
    address = {Cambridge, MA, USA},
    booktitle = {Proceedings of the 29th International Conference on Neural Information Processing Systems - Volume 1},
    pages = {1135–1143},
    numpages = {9},
    location = {Montreal, Canada},
    series = {NIPS'15}
}

@inproceedings{sun2023simple,
  author       = {Mingjie Sun and
                  Zhuang Liu and
                  Anna Bair and
                  J. Zico Kolter},
  title        = {A Simple and Effective Pruning Approach for Large Language Models},
  booktitle    = {The Twelfth International Conference on Learning Representations,
                  {ICLR} 2024, Vienna, Austria, May 7-11, 2024},
  publisher    = {OpenReview.net},
  year         = {2024},
  url          = {https://openreview.net/forum?id=PxoFut3dWW},
  bibsource    = {dblp computer science bibliography, https://dblp.org}
}

@inproceedings{frantar2023sparsegpt,
    author = {Frantar, Elias and Alistarh, Dan},
    title = {SparseGPT: massive language models can be accurately pruned in one-shot},
    year = {2023},
    publisher = {JMLR.org},
    booktitle = {Proceedings of the 40th International Conference on Machine Learning},
    articleno = {414},
    numpages = {15},
    location = {Honolulu, Hawaii, USA},
    series = {ICML'23}
}

@inproceedings{yang2025wanda++,
    title = "Wanda++: Pruning Large Language Models via Regional Gradients",
    author = {Yang, Yifan  and
      Zhen, Kai  and
      Ganesh, Bhavana  and
      Galstyan, Aram  and
      Huybrechts, Goeric  and
      M{\"u}ller, Markus  and
      K{\"u}bler, Jonas M.  and
      Swaminathan, Rupak Vignesh  and
      Mouchtaris, Athanasios  and
      Bodapati, Sravan Babu  and
      Susanj, Nathan  and
      Zhang, Zheng  and
      FitzGerald, Jack  and
      Kumar, Abhishek},
    editor = "Che, Wanxiang  and
      Nabende, Joyce  and
      Shutova, Ekaterina  and
      Pilehvar, Mohammad Taher",
    booktitle = "Findings of the Association for Computational Linguistics: ACL 2025",
    month = jul,
    year = "2025",
    address = "Vienna, Austria",
    publisher = "Association for Computational Linguistics",
    url = "https://aclanthology.org/2025.findings-acl.224/",
    doi = "10.18653/v1/2025.findings-acl.224",
    pages = "4321--4333",
    ISBN = "979-8-89176-256-5"
}

@inproceedings{ashkboos2024slicegpt,
  author       = {Saleh Ashkboos and
                  Maximilian L. Croci and
                  Marcelo Gennari Do Nascimento and
                  Torsten Hoefler and
                  James Hensman},
  title        = {SliceGPT: Compress Large Language Models by Deleting Rows and Columns},
  booktitle    = {The Twelfth International Conference on Learning Representations,
                  {ICLR} 2024, Vienna, Austria, May 7-11, 2024},
  publisher    = {OpenReview.net},
  year         = {2024},
  url          = {https://openreview.net/forum?id=vXxardq6db},
  bibsource    = {dblp computer science bibliography, https://dblp.org}
}

@inproceedings{ma2023llm,
    author = {Ma, Xinyin and Fang, Gongfan and Wang, Xinchao},
    title = {LLM-pruner: on the structural pruning of large language models},
    year = {2023},
    publisher = {Curran Associates Inc.},
    address = {Red Hook, NY, USA},
    booktitle = {Proceedings of the 37th International Conference on Neural Information Processing Systems},
    articleno = {950},
    numpages = {19},
    location = {New Orleans, LA, USA},
    series = {NIPS '23}
}

@inproceedings{yang2024laco,
    title = "{L}a{C}o: Large Language Model Pruning via Layer Collapse",
    author = "Yang, Yifei  and
      Cao, Zouying  and
      Zhao, Hai",
    editor = "Al-Onaizan, Yaser  and
      Bansal, Mohit  and
      Chen, Yun-Nung",
    booktitle = "Findings of the Association for Computational Linguistics: EMNLP 2024",
    month = nov,
    year = "2024",
    address = "Miami, Florida, USA",
    publisher = "Association for Computational Linguistics",
    url = "https://aclanthology.org/2024.findings-emnlp.372/",
    doi = "10.18653/v1/2024.findings-emnlp.372",
    pages = "6401--6417"
}

@inproceedings{an2023fluctuation,
    author = {An, Yongqi and Zhao, Xu and Yu, Tao and Tang, Ming and Wang, Jinqiao},
    title = {Fluctuation-based adaptive structured pruning for large language models},
    year = {2024},
    isbn = {978-1-57735-887-9},
    publisher = {AAAI Press},
    url = {https://doi.org/10.1609/aaai.v38i10.28960},
    doi = {10.1609/aaai.v38i10.28960},
    booktitle = {Proceedings of the Thirty-Eighth AAAI Conference on Artificial Intelligence and Thirty-Sixth Conference on Innovative Applications of Artificial Intelligence and Fourteenth Symposium on Educational Advances in Artificial Intelligence},
    articleno = {1212},
    numpages = {9},
    series = {AAAI'24/IAAI'24/EAAI'24}
}

@misc{yang2025entropy,
  title={Entropy-Based Block Pruning for Efficient Large Language Models}, 
  author={Liangwei Yang and Yuhui Xu and Juntao Tan and Doyen Sahoo and Silvio Savarese and Caiming Xiong and Huan Wang and Shelby Heinecke},
  year={2025},
  eprint={2504.03794},
  archivePrefix={arXiv},
  primaryClass={cs.CL},
  url={https://arxiv.org/abs/2504.03794}, 
}

@misc{gemmateam2025gemma3technicalreport,
    title={Gemma 3 Technical Report}, 
    author={Gemma Team and Aishwarya Kamath and Johan Ferret and Shreya Pathak and Nino Vieillard and Ramona Merhej and Sarah Perrin and Tatiana Matejovicova and Alexandre Ramé and Morgane Rivière and Louis Rouillard and Thomas Mesnard and Geoffrey Cideron and Jean-bastien Grill and Sabela Ramos and Edouard Yvinec and Michelle Casbon and Etienne Pot and Ivo Penchev and Gaël Liu and Francesco Visin and Kathleen Kenealy and Lucas Beyer and Xiaohai Zhai and Anton Tsitsulin and Robert Busa-Fekete and Alex Feng and Noveen Sachdeva and Benjamin Coleman and Yi Gao and Basil Mustafa and Iain Barr and Emilio Parisotto and David Tian and Matan Eyal and Colin Cherry and Jan-Thorsten Peter and Danila Sinopalnikov and Surya Bhupatiraju and Rishabh Agarwal and Mehran Kazemi and Dan Malkin and Ravin Kumar and David Vilar and Idan Brusilovsky and Jiaming Luo and Andreas Steiner and Abe Friesen and Abhanshu Sharma and Abheesht Sharma and Adi Mayrav Gilady and Adrian Goedeckemeyer and Alaa Saade and Alex Feng and Alexander Kolesnikov and Alexei Bendebury and Alvin Abdagic and Amit Vadi and András György and André Susano Pinto and Anil Das and Ankur Bapna and Antoine Miech and Antoine Yang and Antonia Paterson and Ashish Shenoy and Ayan Chakrabarti and Bilal Piot and Bo Wu and Bobak Shahriari and Bryce Petrini and Charlie Chen and Charline Le Lan and Christopher A. Choquette-Choo and CJ Carey and Cormac Brick and Daniel Deutsch and Danielle Eisenbud and Dee Cattle and Derek Cheng and Dimitris Paparas and Divyashree Shivakumar Sreepathihalli and Doug Reid and Dustin Tran and Dustin Zelle and Eric Noland and Erwin Huizenga and Eugene Kharitonov and Frederick Liu and Gagik Amirkhanyan and Glenn Cameron and Hadi Hashemi and Hanna Klimczak-Plucińska and Harman Singh and Harsh Mehta and Harshal Tushar Lehri and Hussein Hazimeh and Ian Ballantyne and Idan Szpektor and Ivan Nardini and Jean Pouget-Abadie and Jetha Chan and Joe Stanton and John Wieting and Jonathan Lai and Jordi Orbay and Joseph Fernandez and Josh Newlan and Ju-yeong Ji and Jyotinder Singh and Kat Black and Kathy Yu and Kevin Hui and Kiran Vodrahalli and Klaus Greff and Linhai Qiu and Marcella Valentine and Marina Coelho and Marvin Ritter and Matt Hoffman and Matthew Watson and Mayank Chaturvedi and Michael Moynihan and Min Ma and Nabila Babar and Natasha Noy and Nathan Byrd and Nick Roy and Nikola Momchev and Nilay Chauhan and Noveen Sachdeva and Oskar Bunyan and Pankil Botarda and Paul Caron and Paul Kishan Rubenstein and Phil Culliton and Philipp Schmid and Pier Giuseppe Sessa and Pingmei Xu and Piotr Stanczyk and Pouya Tafti and Rakesh Shivanna and Renjie Wu and Renke Pan and Reza Rokni and Rob Willoughby and Rohith Vallu and Ryan Mullins and Sammy Jerome and Sara Smoot and Sertan Girgin and Shariq Iqbal and Shashir Reddy and Shruti Sheth and Siim Põder and Sijal Bhatnagar and Sindhu Raghuram Panyam and Sivan Eiger and Susan Zhang and Tianqi Liu and Trevor Yacovone and Tyler Liechty and Uday Kalra and Utku Evci and Vedant Misra and Vincent Roseberry and Vlad Feinberg and Vlad Kolesnikov and Woohyun Han and Woosuk Kwon and Xi Chen and Yinlam Chow and Yuvein Zhu and Zichuan Wei and Zoltan Egyed and Victor Cotruta and Minh Giang and Phoebe Kirk and Anand Rao and Kat Black and Nabila Babar and Jessica Lo and Erica Moreira and Luiz Gustavo Martins and Omar Sanseviero and Lucas Gonzalez and Zach Gleicher and Tris Warkentin and Vahab Mirrokni and Evan Senter and Eli Collins and Joelle Barral and Zoubin Ghahramani and Raia Hadsell and Yossi Matias and D. Sculley and Slav Petrov and Noah Fiedel and Noam Shazeer and Oriol Vinyals and Jeff Dean and Demis Hassabis and Koray Kavukcuoglu and Clement Farabet and Elena Buchatskaya and Jean-Baptiste Alayrac and Rohan Anil and Dmitry and Lepikhin and Sebastian Borgeaud and Olivier Bachem and Armand Joulin and Alek Andreev and Cassidy Hardin and Robert Dadashi and Léonard Hussenot},
    year={2025},
    eprint={2503.19786},
    archivePrefix={arXiv},
    primaryClass={cs.CL},
    url={https://arxiv.org/abs/2503.19786}, 
}

@misc{grattafiori2024llama3herdmodels,
    title={The Llama 3 Herd of Models}, 
    author={Aaron Grattafiori and Abhimanyu Dubey and Abhinav Jauhri and Abhinav Pandey and Abhishek Kadian and Ahmad Al-Dahle and Aiesha Letman and Akhil Mathur and Alan Schelten and Alex Vaughan and Amy Yang and Angela Fan and Anirudh Goyal and Anthony Hartshorn and Aobo Yang and Archi Mitra and Archie Sravankumar and Artem Korenev and Arthur Hinsvark and Arun Rao and Aston Zhang and Aurelien Rodriguez and Austen Gregerson and Ava Spataru and Baptiste Roziere and Bethany Biron and Binh Tang and Bobbie Chern and Charlotte Caucheteux and Chaya Nayak and Chloe Bi and Chris Marra and Chris McConnell and Christian Keller and Christophe Touret and Chunyang Wu and Corinne Wong and Cristian Canton Ferrer and Cyrus Nikolaidis and Damien Allonsius and Daniel Song and Danielle Pintz and Danny Livshits and Danny Wyatt and David Esiobu and Dhruv Choudhary and Dhruv Mahajan and Diego Garcia-Olano and Diego Perino and Dieuwke Hupkes and Egor Lakomkin and Ehab AlBadawy and Elina Lobanova and Emily Dinan and Eric Michael Smith and Filip Radenovic and Francisco Guzmán and Frank Zhang and Gabriel Synnaeve and Gabrielle Lee and Georgia Lewis Anderson and Govind Thattai and Graeme Nail and Gregoire Mialon and Guan Pang and Guillem Cucurell and Hailey Nguyen and Hannah Korevaar and Hu Xu and Hugo Touvron and Iliyan Zarov and Imanol Arrieta Ibarra and Isabel Kloumann and Ishan Misra and Ivan Evtimov and Jack Zhang and Jade Copet and Jaewon Lee and Jan Geffert and Jana Vranes and Jason Park and Jay Mahadeokar and Jeet Shah and Jelmer van der Linde and Jennifer Billock and Jenny Hong and Jenya Lee and Jeremy Fu and Jianfeng Chi and Jianyu Huang and Jiawen Liu and Jie Wang and Jiecao Yu and Joanna Bitton and Joe Spisak and Jongsoo Park and Joseph Rocca and Joshua Johnstun and Joshua Saxe and Junteng Jia and Kalyan Vasuden Alwala and Karthik Prasad and Kartikeya Upasani and Kate Plawiak and Ke Li and Kenneth Heafield and Kevin Stone and Khalid El-Arini and Krithika Iyer and Kshitiz Malik and Kuenley Chiu and Kunal Bhalla and Kushal Lakhotia and Lauren Rantala-Yeary and Laurens van der Maaten and Lawrence Chen and Liang Tan and Liz Jenkins and Louis Martin and Lovish Madaan and Lubo Malo and Lukas Blecher and Lukas Landzaat and Luke de Oliveira and Madeline Muzzi and Mahesh Pasupuleti and Mannat Singh and Manohar Paluri and Marcin Kardas and Maria Tsimpoukelli and Mathew Oldham and Mathieu Rita and Maya Pavlova and Melanie Kambadur and Mike Lewis and Min Si and Mitesh Kumar Singh and Mona Hassan and Naman Goyal and Narjes Torabi and Nikolay Bashlykov and Nikolay Bogoychev and Niladri Chatterji and Ning Zhang and Olivier Duchenne and Onur Çelebi and Patrick Alrassy and Pengchuan Zhang and Pengwei Li and Petar Vasic and Peter Weng and Prajjwal Bhargava and Pratik Dubal and Praveen Krishnan and Punit Singh Koura and Puxin Xu and Qing He and Qingxiao Dong and Ragavan Srinivasan and Raj Ganapathy and Ramon Calderer and Ricardo Silveira Cabral and Robert Stojnic and Roberta Raileanu and Rohan Maheswari and Rohit Girdhar and Rohit Patel and Romain Sauvestre and Ronnie Polidoro and Roshan Sumbaly and Ross Taylor and Ruan Silva and Rui Hou and Rui Wang and Saghar Hosseini and Sahana Chennabasappa and Sanjay Singh and Sean Bell and Seohyun Sonia Kim and Sergey Edunov and Shaoliang Nie and Sharan Narang and Sharath Raparthy and Sheng Shen and Shengye Wan and Shruti Bhosale and Shun Zhang and Simon Vandenhende and Soumya Batra and Spencer Whitman and Sten Sootla and Stephane Collot and Suchin Gururangan and Sydney Borodinsky and Tamar Herman and Tara Fowler and Tarek Sheasha and Thomas Georgiou and Thomas Scialom and Tobias Speckbacher and Todor Mihaylov and Tong Xiao and Ujjwal Karn and Vedanuj Goswami and Vibhor Gupta and Vignesh Ramanathan and Viktor Kerkez and Vincent Gonguet and Virginie Do and Vish Vogeti and Vítor Albiero and Vladan Petrovic and Weiwei Chu and Wenhan Xiong and Wenyin Fu and Whitney Meers and Xavier Martinet and Xiaodong Wang and Xiaofang Wang and Xiaoqing Ellen Tan and Xide Xia and Xinfeng Xie and Xuchao Jia and Xuewei Wang and Yaelle Goldschlag and Yashesh Gaur and Yasmine Babaei and Yi Wen and Yiwen Song and Yuchen Zhang and Yue Li and Yuning Mao and Zacharie Delpierre Coudert and Zheng Yan and Zhengxing Chen and Zoe Papakipos and Aaditya Singh and Aayushi Srivastava and Abha Jain and Adam Kelsey and Adam Shajnfeld and Adithya Gangidi and Adolfo Victoria and Ahuva Goldstand and Ajay Menon and Ajay Sharma and Alex Boesenberg and Alexei Baevski and Allie Feinstein and Amanda Kallet and Amit Sangani and Amos Teo and Anam Yunus and Andrei Lupu and Andres Alvarado and Andrew Caples and Andrew Gu and Andrew Ho and Andrew Poulton and Andrew Ryan and Ankit Ramchandani and Annie Dong and Annie Franco and Anuj Goyal and Aparajita Saraf and Arkabandhu Chowdhury and Ashley Gabriel and Ashwin Bharambe and Assaf Eisenman and Azadeh Yazdan and Beau James and Ben Maurer and Benjamin Leonhardi and Bernie Huang and Beth Loyd and Beto De Paola and Bhargavi Paranjape and Bing Liu and Bo Wu and Boyu Ni and Braden Hancock and Bram Wasti and Brandon Spence and Brani Stojkovic and Brian Gamido and Britt Montalvo and Carl Parker and Carly Burton and Catalina Mejia and Ce Liu and Changhan Wang and Changkyu Kim and Chao Zhou and Chester Hu and Ching-Hsiang Chu and Chris Cai and Chris Tindal and Christoph Feichtenhofer and Cynthia Gao and Damon Civin and Dana Beaty and Daniel Kreymer and Daniel Li and David Adkins and David Xu and Davide Testuggine and Delia David and Devi Parikh and Diana Liskovich and Didem Foss and Dingkang Wang and Duc Le and Dustin Holland and Edward Dowling and Eissa Jamil and Elaine Montgomery and Eleonora Presani and Emily Hahn and Emily Wood and Eric-Tuan Le and Erik Brinkman and Esteban Arcaute and Evan Dunbar and Evan Smothers and Fei Sun and Felix Kreuk and Feng Tian and Filippos Kokkinos and Firat Ozgenel and Francesco Caggioni and Frank Kanayet and Frank Seide and Gabriela Medina Florez and Gabriella Schwarz and Gada Badeer and Georgia Swee and Gil Halpern and Grant Herman and Grigory Sizov and Guangyi and Zhang and Guna Lakshminarayanan and Hakan Inan and Hamid Shojanazeri and Han Zou and Hannah Wang and Hanwen Zha and Haroun Habeeb and Harrison Rudolph and Helen Suk and Henry Aspegren and Hunter Goldman and Hongyuan Zhan and Ibrahim Damlaj and Igor Molybog and Igor Tufanov and Ilias Leontiadis and Irina-Elena Veliche and Itai Gat and Jake Weissman and James Geboski and James Kohli and Janice Lam and Japhet Asher and Jean-Baptiste Gaya and Jeff Marcus and Jeff Tang and Jennifer Chan and Jenny Zhen and Jeremy Reizenstein and Jeremy Teboul and Jessica Zhong and Jian Jin and Jingyi Yang and Joe Cummings and Jon Carvill and Jon Shepard and Jonathan McPhie and Jonathan Torres and Josh Ginsburg and Junjie Wang and Kai Wu and Kam Hou U and Karan Saxena and Kartikay Khandelwal and Katayoun Zand and Kathy Matosich and Kaushik Veeraraghavan and Kelly Michelena and Keqian Li and Kiran Jagadeesh and Kun Huang and Kunal Chawla and Kyle Huang and Lailin Chen and Lakshya Garg and Lavender A and Leandro Silva and Lee Bell and Lei Zhang and Liangpeng Guo and Licheng Yu and Liron Moshkovich and Luca Wehrstedt and Madian Khabsa and Manav Avalani and Manish Bhatt and Martynas Mankus and Matan Hasson and Matthew Lennie and Matthias Reso and Maxim Groshev and Maxim Naumov and Maya Lathi and Meghan Keneally and Miao Liu and Michael L. Seltzer and Michal Valko and Michelle Restrepo and Mihir Patel and Mik Vyatskov and Mikayel Samvelyan and Mike Clark and Mike Macey and Mike Wang and Miquel Jubert Hermoso and Mo Metanat and Mohammad Rastegari and Munish Bansal and Nandhini Santhanam and Natascha Parks and Natasha White and Navyata Bawa and Nayan Singhal and Nick Egebo and Nicolas Usunier and Nikhil Mehta and Nikolay Pavlovich Laptev and Ning Dong and Norman Cheng and Oleg Chernoguz and Olivia Hart and Omkar Salpekar and Ozlem Kalinli and Parkin Kent and Parth Parekh and Paul Saab and Pavan Balaji and Pedro Rittner and Philip Bontrager and Pierre Roux and Piotr Dollar and Polina Zvyagina and Prashant Ratanchandani and Pritish Yuvraj and Qian Liang and Rachad Alao and Rachel Rodriguez and Rafi Ayub and Raghotham Murthy and Raghu Nayani and Rahul Mitra and Rangaprabhu Parthasarathy and Raymond Li and Rebekkah Hogan and Robin Battey and Rocky Wang and Russ Howes and Ruty Rinott and Sachin Mehta and Sachin Siby and Sai Jayesh Bondu and Samyak Datta and Sara Chugh and Sara Hunt and Sargun Dhillon and Sasha Sidorov and Satadru Pan and Saurabh Mahajan and Saurabh Verma and Seiji Yamamoto and Sharadh Ramaswamy and Shaun Lindsay and Shaun Lindsay and Sheng Feng and Shenghao Lin and Shengxin Cindy Zha and Shishir Patil and Shiva Shankar and Shuqiang Zhang and Shuqiang Zhang and Sinong Wang and Sneha Agarwal and Soji Sajuyigbe and Soumith Chintala and Stephanie Max and Stephen Chen and Steve Kehoe and Steve Satterfield and Sudarshan Govindaprasad and Sumit Gupta and Summer Deng and Sungmin Cho and Sunny Virk and Suraj Subramanian and Sy Choudhury and Sydney Goldman and Tal Remez and Tamar Glaser and Tamara Best and Thilo Koehler and Thomas Robinson and Tianhe Li and Tianjun Zhang and Tim Matthews and Timothy Chou and Tzook Shaked and Varun Vontimitta and Victoria Ajayi and Victoria Montanez and Vijai Mohan and Vinay Satish Kumar and Vishal Mangla and Vlad Ionescu and Vlad Poenaru and Vlad Tiberiu Mihailescu and Vladimir Ivanov and Wei Li and Wenchen Wang and Wenwen Jiang and Wes Bouaziz and Will Constable and Xiaocheng Tang and Xiaojian Wu and Xiaolan Wang and Xilun Wu and Xinbo Gao and Yaniv Kleinman and Yanjun Chen and Ye Hu and Ye Jia and Ye Qi and Yenda Li and Yilin Zhang and Ying Zhang and Yossi Adi and Youngjin Nam and Yu and Wang and Yu Zhao and Yuchen Hao and Yundi Qian and Yunlu Li and Yuzi He and Zach Rait and Zachary DeVito and Zef Rosnbrick and Zhaoduo Wen and Zhenyu Yang and Zhiwei Zhao and Zhiyu Ma},
    year={2024},
    eprint={2407.21783},
    archivePrefix={arXiv},
    primaryClass={cs.AI},
    url={https://arxiv.org/abs/2407.21783}, 
}

@misc{yang2025qwen3technicalreport,
      title={Qwen3 Technical Report}, 
      author={An Yang and Anfeng Li and Baosong Yang and Beichen Zhang and Binyuan Hui and Bo Zheng and Bowen Yu and Chang Gao and Chengen Huang and Chenxu Lv and Chujie Zheng and Dayiheng Liu and Fan Zhou and Fei Huang and Feng Hu and Hao Ge and Haoran Wei and Huan Lin and Jialong Tang and Jian Yang and Jianhong Tu and Jianwei Zhang and Jianxin Yang and Jiaxi Yang and Jing Zhou and Jingren Zhou and Junyang Lin and Kai Dang and Keqin Bao and Kexin Yang and Le Yu and Lianghao Deng and Mei Li and Mingfeng Xue and Mingze Li and Pei Zhang and Peng Wang and Qin Zhu and Rui Men and Ruize Gao and Shixuan Liu and Shuang Luo and Tianhao Li and Tianyi Tang and Wenbiao Yin and Xingzhang Ren and Xinyu Wang and Xinyu Zhang and Xuancheng Ren and Yang Fan and Yang Su and Yichang Zhang and Yinger Zhang and Yu Wan and Yuqiong Liu and Zekun Wang and Zeyu Cui and Zhenru Zhang and Zhipeng Zhou and Zihan Qiu},
      year={2025},
      eprint={2505.09388},
      archivePrefix={arXiv},
      primaryClass={cs.CL},
      url={https://arxiv.org/abs/2505.09388}, 
}

@software{lintang_sutawika_2023_10256836,
  author       = {Gao, Leo and Tow, Jonathan and Abbasi, Baber and Biderman, Stella and Black, Sid and DiPofi, Anthony and Foster, Charles and Golding, Laurence and Hsu, Jeffrey and Le Noac'h, Alain and Li, Haonan and McDonell, Kyle and Muennighoff, Niklas and Ociepa, Chris and Phang, Jason and Reynolds, Laria and Schoelkopf, Hailey and Skowron, Aviya and Sutawika, Lintang and Tang, Eric and Thite, Anish and Wang, Ben and Wang, Kevin and Zou, Andy},
  title        = {The Language Model Evaluation Harness},
  month        = 07,
  year         = 2024,
  publisher    = {Zenodo},
  version      = {v0.4.3},
  doi          = {10.5281/zenodo.12608602},
  url          = {https://zenodo.org/records/12608602}
}

@inproceedings{merity2017pointer,
    author       = {Stephen Merity and
                  Caiming Xiong and
                  James Bradbury and
                  Richard Socher},
    title        = {Pointer Sentinel Mixture Models},
    booktitle    = {5th International Conference on Learning Representations, {ICLR} 2017,
                  Toulon, France, April 24-26, 2017, Conference Track Proceedings},
    publisher    = {OpenReview.net},
    year         = {2017},
    url          = {https://openreview.net/forum?id=Byj72udxe},
    bibsource    = {dblp computer science bibliography, https://dblp.org}
}

@misc{Clark2018ThinkYH,
    title={Think you have Solved Question Answering? Try ARC, the AI2 Reasoning Challenge}, 
    author={Peter Clark and Isaac Cowhey and Oren Etzioni and Tushar Khot and Ashish Sabharwal and Carissa Schoenick and Oyvind Tafjord},
    year={2018},
    eprint={1803.05457},
    archivePrefix={arXiv},
    primaryClass={cs.AI},
    url={https://arxiv.org/abs/1803.05457}, 
}

@inproceedings{clark-etal-2019-boolq,
    title = "{B}ool{Q}: Exploring the Surprising Difficulty of Natural Yes/No Questions",
    author = "Clark, Christopher  and
      Lee, Kenton  and
      Chang, Ming-Wei  and
      Kwiatkowski, Tom  and
      Collins, Michael  and
      Toutanova, Kristina",
    editor = "Burstein, Jill  and
      Doran, Christy  and
      Solorio, Thamar",
    booktitle = "Proceedings of the 2019 Conference of the North {A}merican Chapter of the Association for Computational Linguistics: Human Language Technologies, Volume 1 (Long and Short Papers)",
    month = jun,
    year = "2019",
    address = "Minneapolis, Minnesota",
    publisher = "Association for Computational Linguistics",
    url = "https://aclanthology.org/N19-1300/",
    doi = "10.18653/v1/N19-1300",
    pages = "2924--2936"
}

@inproceedings{mihaylov-etal-2018-suit,
    title = "Can a Suit of Armor Conduct Electricity? A New Dataset for Open Book Question Answering",
    author = "Mihaylov, Todor  and
      Clark, Peter  and
      Khot, Tushar  and
      Sabharwal, Ashish",
    editor = "Riloff, Ellen  and
      Chiang, David  and
      Hockenmaier, Julia  and
      Tsujii, Jun{'}ichi",
    booktitle = "Proceedings of the 2018 Conference on Empirical Methods in Natural Language Processing",
    month = oct # "-" # nov,
    year = "2018",
    address = "Brussels, Belgium",
    publisher = "Association for Computational Linguistics",
    url = "https://aclanthology.org/D18-1260/",
    doi = "10.18653/v1/D18-1260",
    pages = "2381--2391"
}

@inproceedings{DBLP:conf/aaai/BiskZLGC20,
  author       = {Yonatan Bisk and
                  Rowan Zellers and
                  Ronan Le Bras and
                  Jianfeng Gao and
                  Yejin Choi},
  title        = {{PIQA:} Reasoning about Physical Commonsense in Natural Language},
  booktitle    = {The Thirty-Fourth {AAAI} Conference on Artificial Intelligence, {AAAI}
                  2020, The Thirty-Second Innovative Applications of Artificial Intelligence
                  Conference, {IAAI} 2020, The Tenth {AAAI} Symposium on Educational
                  Advances in Artificial Intelligence, {EAAI} 2020, New York, NY, USA,
                  February 7-12, 2020},
  pages        = {7432--7439},
  publisher    = {{AAAI} Press},
  year         = {2020},
  url          = {https://doi.org/10.1609/aaai.v34i05.6239},
  doi          = {10.1609/AAAI.V34I05.6239},
  bibsource    = {dblp computer science bibliography, https://dblp.org}
}

@article{sakaguchi2019winogrande,
    author = {Sakaguchi, Keisuke and Bras, Ronan Le and Bhagavatula, Chandra and Choi, Yejin},
    title = {WinoGrande: an adversarial winograd schema challenge at scale},
    year = {2021},
    issue_date = {September 2021},
    publisher = {Association for Computing Machinery},
    address = {New York, NY, USA},
    volume = {64},
    number = {9},
    issn = {0001-0782},
    url = {https://doi.org/10.1145/3474381},
    doi = {10.1145/3474381},
    journal = {Commun. ACM},
    month = aug,
    pages = {99–106},
    numpages = {8}
}

@inproceedings{zellers2019hellaswag,
    title = "{H}ella{S}wag: Can a Machine Really Finish Your Sentence?",
    author = "Zellers, Rowan  and
      Holtzman, Ari  and
      Bisk, Yonatan  and
      Farhadi, Ali  and
      Choi, Yejin",
    editor = "Korhonen, Anna  and
      Traum, David  and
      M{\`a}rquez, Llu{\'i}s",
    booktitle = "Proceedings of the 57th Annual Meeting of the Association for Computational Linguistics",
    month = jul,
    year = "2019",
    address = "Florence, Italy",
    publisher = "Association for Computational Linguistics",
    url = "https://aclanthology.org/P19-1472/",
    doi = "10.18653/v1/P19-1472",
    pages = "4791--4800"
}

@inproceedings{hendrycks2021measuring,
    author       = {Dan Hendrycks and
                  Collin Burns and
                  Steven Basart and
                  Andy Zou and
                  Mantas Mazeika and
                  Dawn Song and
                  Jacob Steinhardt},
    title        = {Measuring Massive Multitask Language Understanding},
    booktitle    = {9th International Conference on Learning Representations, {ICLR} 2021,
                  Virtual Event, Austria, May 3-7, 2021},
    publisher    = {OpenReview.net},
    year         = {2021},
    url          = {https://openreview.net/forum?id=d7KBjmI3GmQ},
    bibsource    = {dblp computer science bibliography, https://dblp.org}
}

@misc{chen2021evaluating,
    title={Evaluating Large Language Models Trained on Code}, 
    author={Mark Chen and Jerry Tworek and Heewoo Jun and Qiming Yuan and Henrique Ponde de Oliveira Pinto and Jared Kaplan and Harri Edwards and Yuri Burda and Nicholas Joseph and Greg Brockman and Alex Ray and Raul Puri and Gretchen Krueger and Michael Petrov and Heidy Khlaaf and Girish Sastry and Pamela Mishkin and Brooke Chan and Scott Gray and Nick Ryder and Mikhail Pavlov and Alethea Power and Lukasz Kaiser and Mohammad Bavarian and Clemens Winter and Philippe Tillet and Felipe Petroski Such and Dave Cummings and Matthias Plappert and Fotios Chantzis and Elizabeth Barnes and Ariel Herbert-Voss and William Hebgen Guss and Alex Nichol and Alex Paino and Nikolas Tezak and Jie Tang and Igor Babuschkin and Suchir Balaji and Shantanu Jain and William Saunders and Christopher Hesse and Andrew N. Carr and Jan Leike and Josh Achiam and Vedant Misra and Evan Morikawa and Alec Radford and Matthew Knight and Miles Brundage and Mira Murati and Katie Mayer and Peter Welinder and Bob McGrew and Dario Amodei and Sam McCandlish and Ilya Sutskever and Wojciech Zaremba},
    year={2021},
    eprint={2107.03374},
    archivePrefix={arXiv},
    primaryClass={cs.LG},
    url={https://arxiv.org/abs/2107.03374}, 
}

@inproceedings{joshi2017triviaqa,
    title = "{T}rivia{QA}: A Large Scale Distantly Supervised Challenge Dataset for Reading Comprehension",
    author = "Joshi, Mandar  and
      Choi, Eunsol  and
      Weld, Daniel  and
      Zettlemoyer, Luke",
    editor = "Barzilay, Regina  and
      Kan, Min-Yen",
    booktitle = "Proceedings of the 55th Annual Meeting of the Association for Computational Linguistics (Volume 1: Long Papers)",
    month = jul,
    year = "2017",
    address = "Vancouver, Canada",
    publisher = "Association for Computational Linguistics",
    url = "https://aclanthology.org/P17-1147/",
    doi = "10.18653/v1/P17-1147",
    pages = "1601--1611"
}

@misc{cobbe2021training,
    title={Training Verifiers to Solve Math Word Problems}, 
    author={Karl Cobbe and Vineet Kosaraju and Mohammad Bavarian and Mark Chen and Heewoo Jun and Lukasz Kaiser and Matthias Plappert and Jerry Tworek and Jacob Hilton and Reiichiro Nakano and Christopher Hesse and John Schulman},
    year={2021},
    eprint={2110.14168},
    archivePrefix={arXiv},
    primaryClass={cs.LG},
    url={https://arxiv.org/abs/2110.14168}, 
}

@inproceedings{yang2018hotpotqa,
    title = "{H}otpot{QA}: A Dataset for Diverse, Explainable Multi-hop Question Answering",
    author = "Yang, Zhilin  and
      Qi, Peng  and
      Zhang, Saizheng  and
      Bengio, Yoshua  and
      Cohen, William  and
      Salakhutdinov, Ruslan  and
      Manning, Christopher D.",
    editor = "Riloff, Ellen  and
      Chiang, David  and
      Hockenmaier, Julia  and
      Tsujii, Jun{'}ichi",
    booktitle = "Proceedings of the 2018 Conference on Empirical Methods in Natural Language Processing",
    month = oct # "-" # nov,
    year = "2018",
    address = "Brussels, Belgium",
    publisher = "Association for Computational Linguistics",
    url = "https://aclanthology.org/D18-1259/",
    doi = "10.18653/v1/D18-1259",
    pages = "2369--2380"
}

@inproceedings{bai2023longbench,
  title = "{L}ong{B}ench: A Bilingual, Multitask Benchmark for Long Context Understanding",
    author = "Bai, Yushi  and
      Lv, Xin  and
      Zhang, Jiajie  and
      Lyu, Hongchang  and
      Tang, Jiankai  and
      Huang, Zhidian  and
      Du, Zhengxiao  and
      Liu, Xiao  and
      Zeng, Aohan  and
      Hou, Lei  and
      Dong, Yuxiao  and
      Tang, Jie  and
      Li, Juanzi",
    editor = "Ku, Lun-Wei  and
      Martins, Andre  and
      Srikumar, Vivek",
    booktitle = "Proceedings of the 62nd Annual Meeting of the Association for Computational Linguistics (Volume 1: Long Papers)",
    month = aug,
    year = "2024",
    address = "Bangkok, Thailand",
    publisher = "Association for Computational Linguistics",
    url = "https://aclanthology.org/2024.acl-long.172/",
    doi = "10.18653/v1/2024.acl-long.172",
    pages = "3119--3137"
}

@inproceedings{dasigi2021qasper,
    title = "A Dataset of Information-Seeking Questions and Answers Anchored in Research Papers",
    author = "Dasigi, Pradeep  and
      Lo, Kyle  and
      Beltagy, Iz  and
      Cohan, Arman  and
      Smith, Noah A.  and
      Gardner, Matt",
    editor = "Toutanova, Kristina  and
      Rumshisky, Anna  and
      Zettlemoyer, Luke  and
      Hakkani-Tur, Dilek  and
      Beltagy, Iz  and
      Bethard, Steven  and
      Cotterell, Ryan  and
      Chakraborty, Tanmoy  and
      Zhou, Yichao",
    booktitle = "Proceedings of the 2021 Conference of the North American Chapter of the Association for Computational Linguistics: Human Language Technologies",
    month = jun,
    year = "2021",
    address = "Online",
    publisher = "Association for Computational Linguistics",
    url = "https://aclanthology.org/2021.naacl-main.365/",
    doi = "10.18653/v1/2021.naacl-main.365",
    pages = "4599--4610"
}

@inproceedings{huang2021efficient,
    title = "Efficient Attentions for Long Document Summarization",
    author = "Huang, Luyang  and
      Cao, Shuyang  and
      Parulian, Nikolaus  and
      Ji, Heng  and
      Wang, Lu",
    editor = "Toutanova, Kristina  and
      Rumshisky, Anna  and
      Zettlemoyer, Luke  and
      Hakkani-Tur, Dilek  and
      Beltagy, Iz  and
      Bethard, Steven  and
      Cotterell, Ryan  and
      Chakraborty, Tanmoy  and
      Zhou, Yichao",
    booktitle = "Proceedings of the 2021 Conference of the North American Chapter of the Association for Computational Linguistics: Human Language Technologies",
    month = jun,
    year = "2021",
    address = "Online",
    publisher = "Association for Computational Linguistics",
    url = "https://aclanthology.org/2021.naacl-main.112/",
    doi = "10.18653/v1/2021.naacl-main.112",
    pages = "1419--1436"
}

@misc{miller2023attention,
  author = {Miller, Evan},
  title = {Attention Is Off By One},
  year = {2023},
  month = {July},
  day = {24},
  url = {https://www.evanmiller.org/attention-is-off-by-one.html}
}

\clearpage
\twocolumn[{
    \begin{center}
        \vspace{2em} 
        \includegraphics[width=0.24\linewidth]{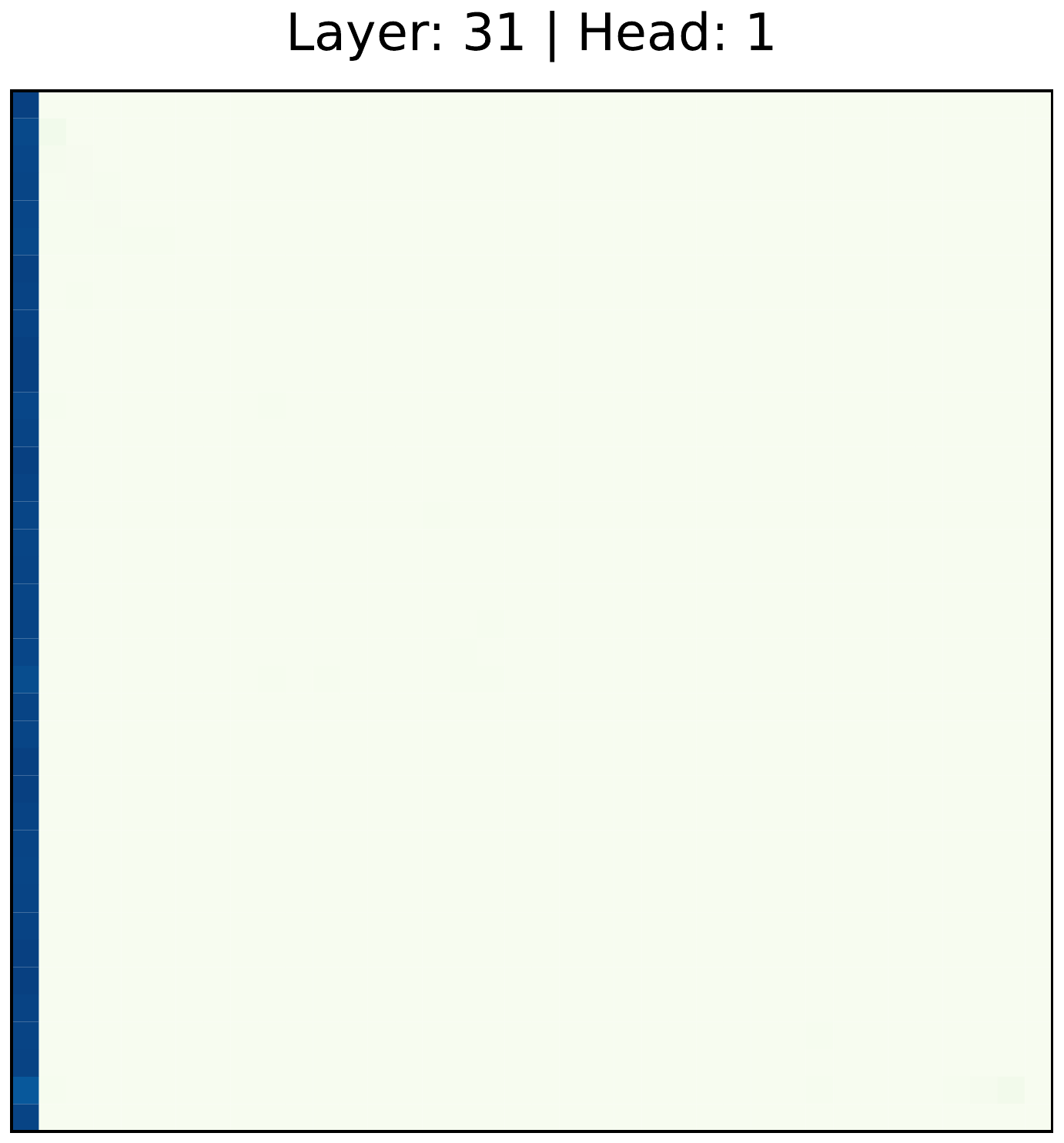}
        \includegraphics[width=0.24\linewidth]{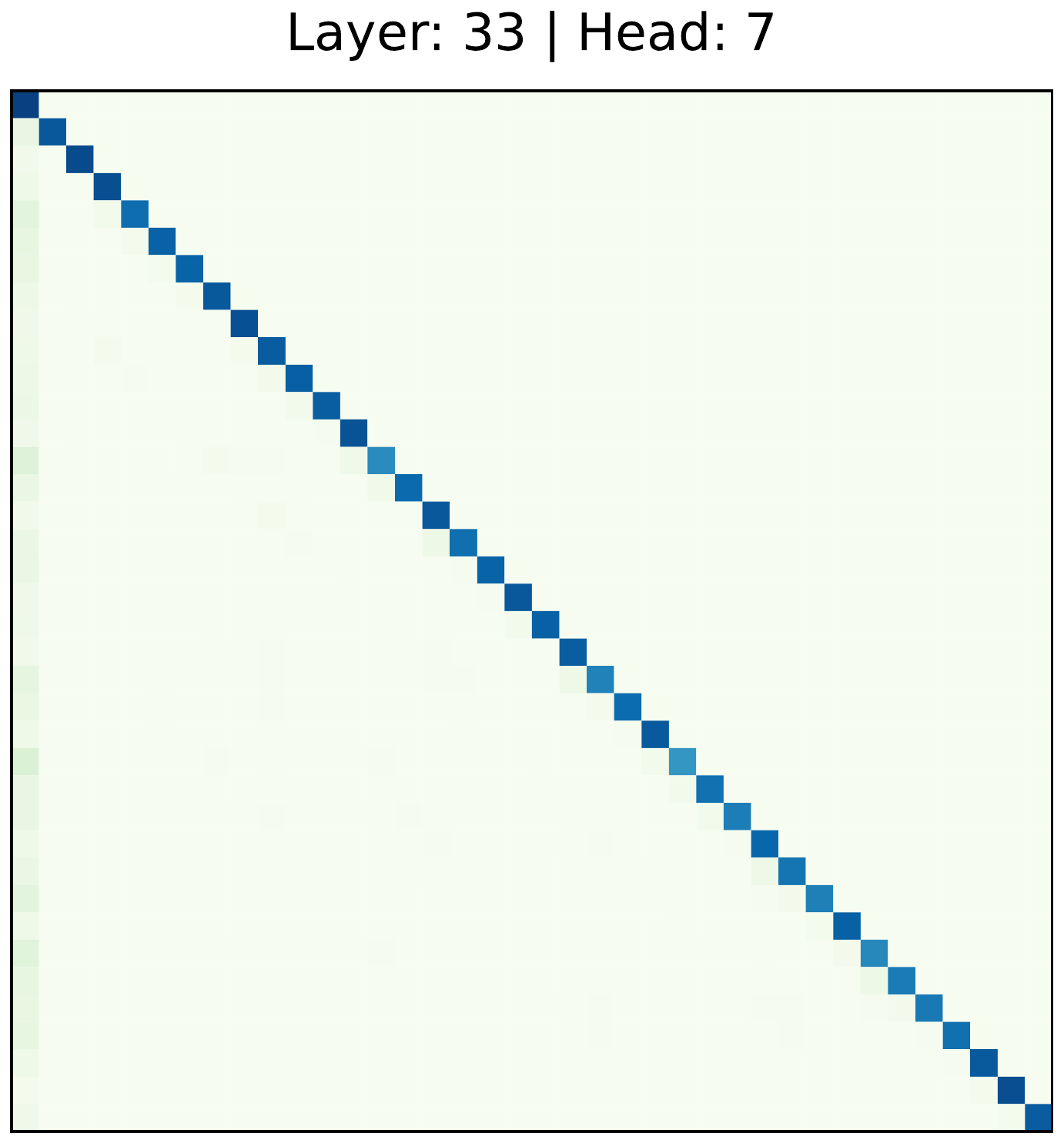}
        \includegraphics[width=0.24\linewidth]{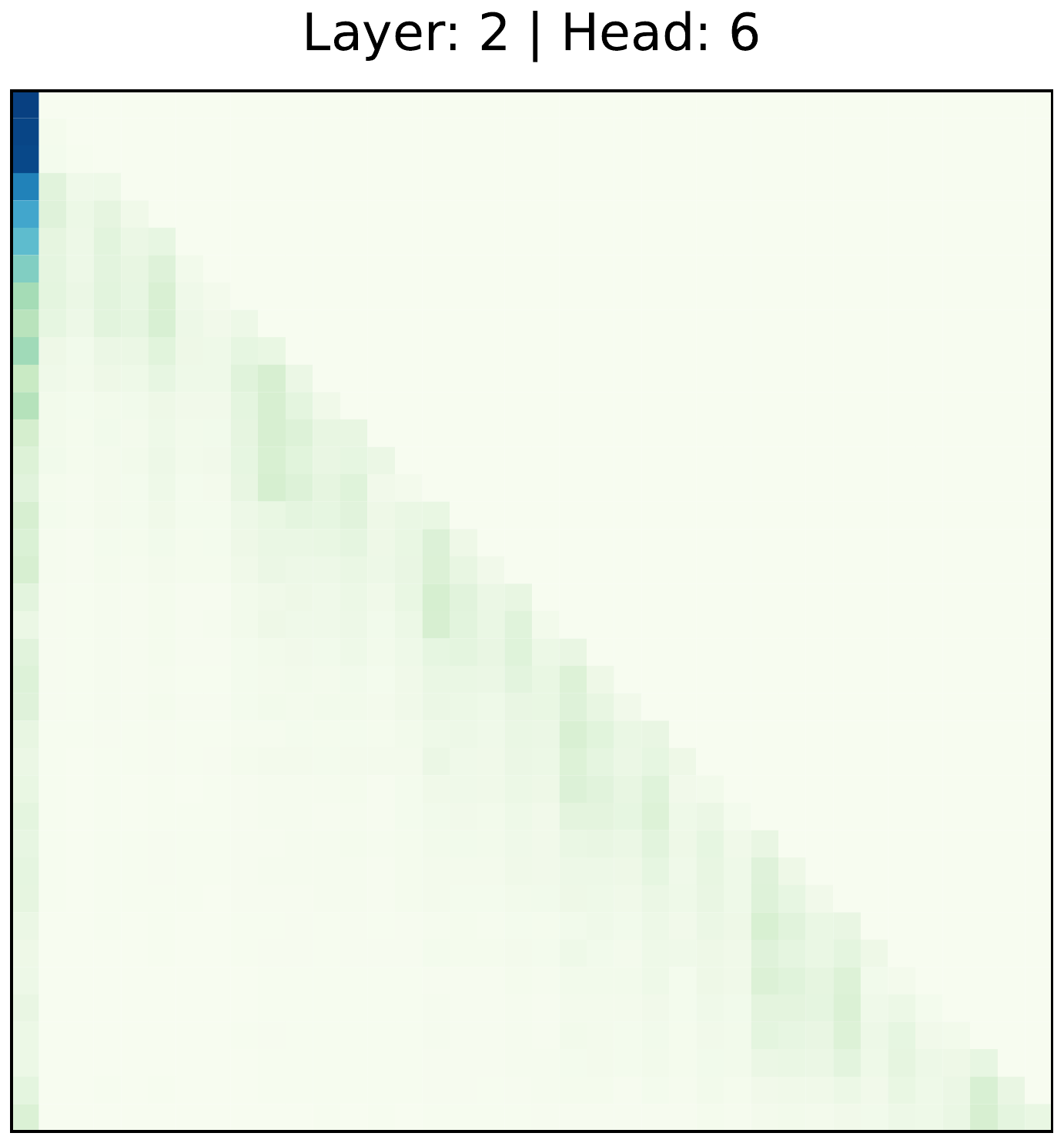}
        \includegraphics[width=0.24\linewidth]{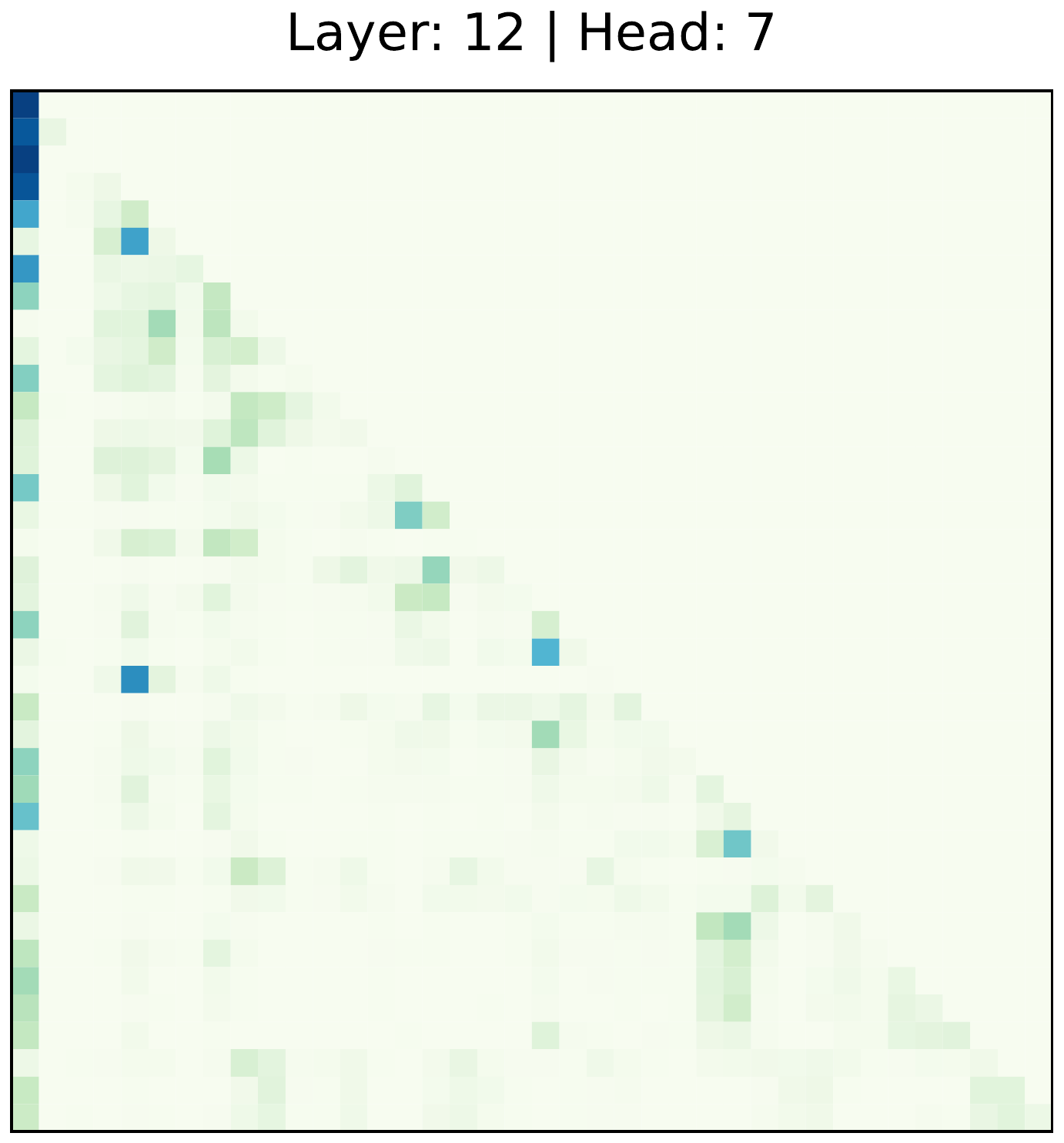}

        \captionof{figure}{Visualization of four representative attention patterns in Gemma-3-4B during MMLU tasks. From left to right, the maps illustrate: (1) \textbf{\BOS Sink} showing intense focus on the initial token (Layer 31, Head 1); (2) \textbf{Diagonal (Self-Attention)} with weights concentrated strictly on the query token (Layer 33, Head 7); (3) \textbf{Uniform} broad attention distribution primarily observed in lower layers (Layer 2, Head 6); and (4) \textbf{Random} sparse attention patterns (Layer 12, Head 7).}
        \label{fig:attention_patterns}
        \vspace{2em} 
    \end{center}
}]
\appendix

\section{Patterns of Attention}

\label{sec:attention_patterns}

To qualitatively characterize the attention behaviors underlying model redundancy, we visualize the attention maps of Gemma-3-4B during inference. As illustrated in Figure~\ref{fig:attention_patterns}, we identify four representative patterns that define the functional landscape of the attention mechanism: \BOS Sink, Diagonal, Uniform, and Random. Our qualitative analysis of these signatures yields three primary findings regarding the structural properties of these attention heads.

\paragraph{Dominance of Non-Transformative Patterns} A significant majority of attention heads exhibit either \BOS sink or Diagonal patterns. While \BOS sinks are known as repositories for unnecessary attention weights, we hypothesize that Diagonal patterns may serve a similar function. In a residual architecture, a head attending solely to its own position effectively acts as an identity map, contributing no new contextual information to the hidden state and thus remaining functionally redundant.

\paragraph{Layer-wise Specialization} Uniform patterns, which facilitate broad information mixing, are predominantly concentrated in the lower layers (e.g., Layer 2). This suggests that global contextual integration occurs early in the forward pass, while deeper layers transition toward specialized \emph{dumping grounds} behaviors like \BOS or Diagonal sinks.

\begin{figure*}[p]
    \centering
    \includegraphics[width=1.0\linewidth]{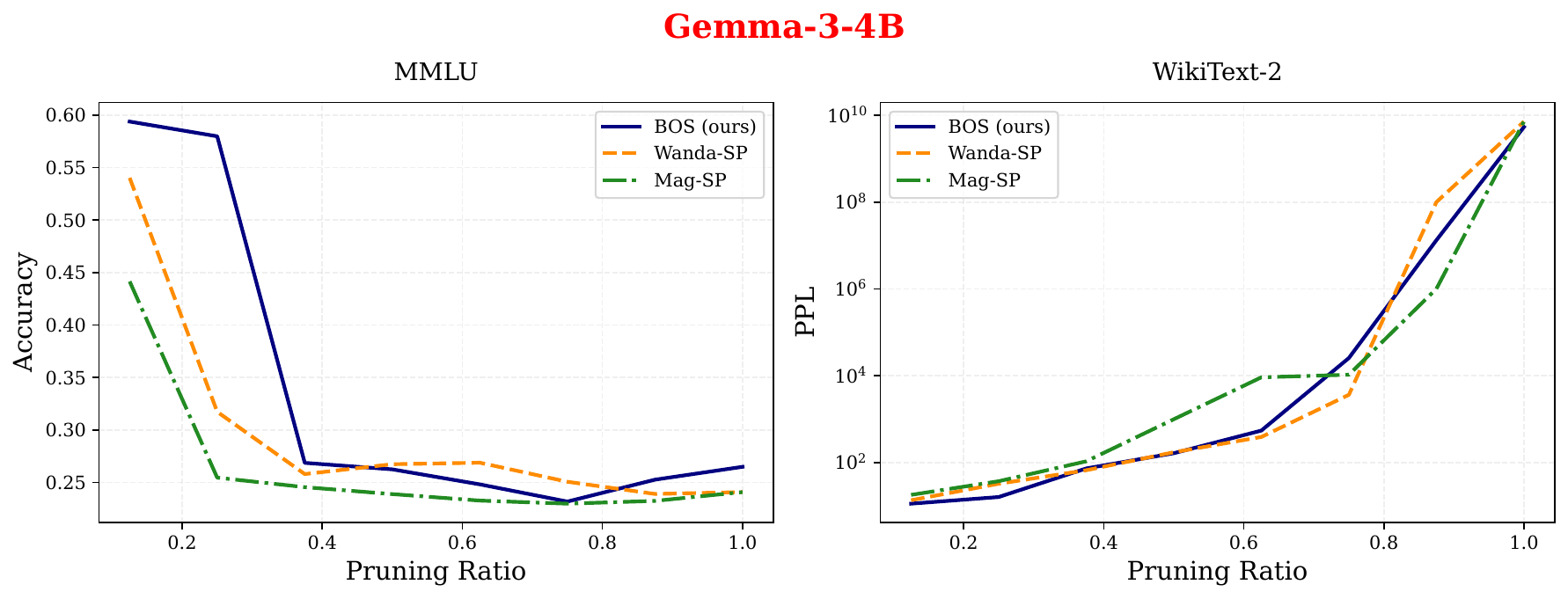}
    \includegraphics[width=1.0\linewidth]{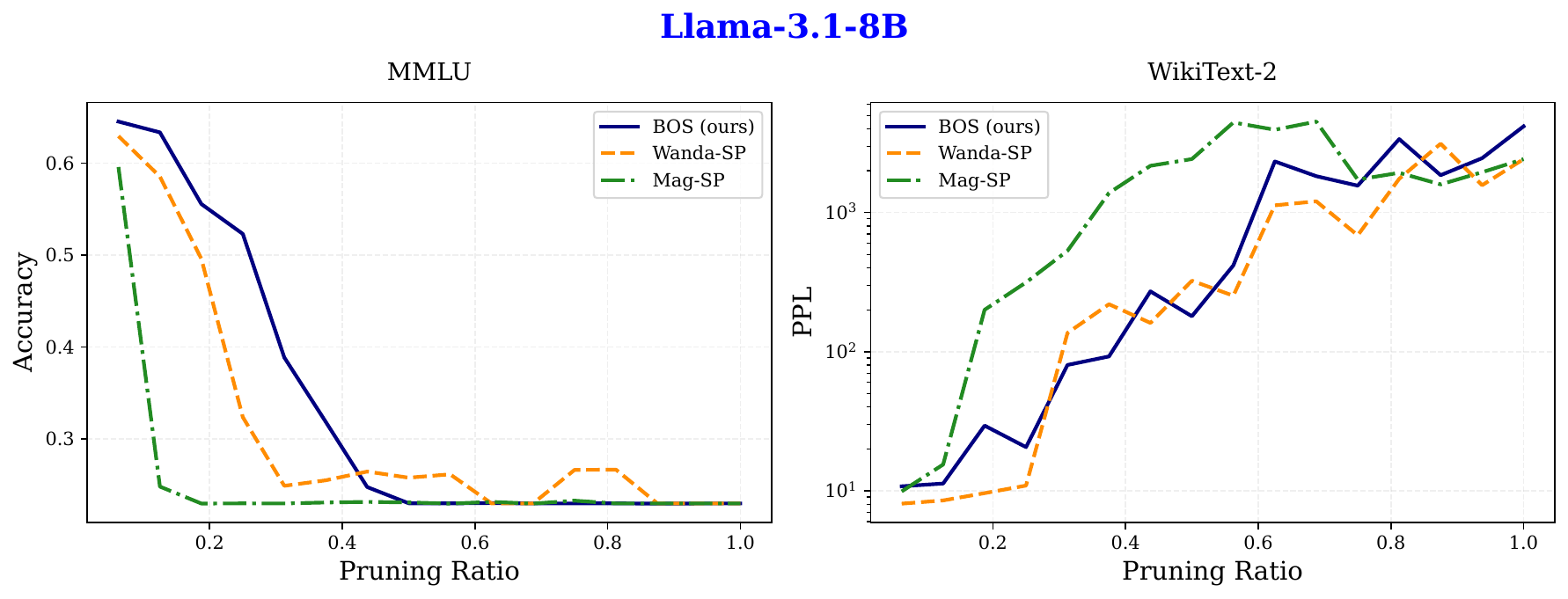}
    \includegraphics[width=1.0\linewidth]{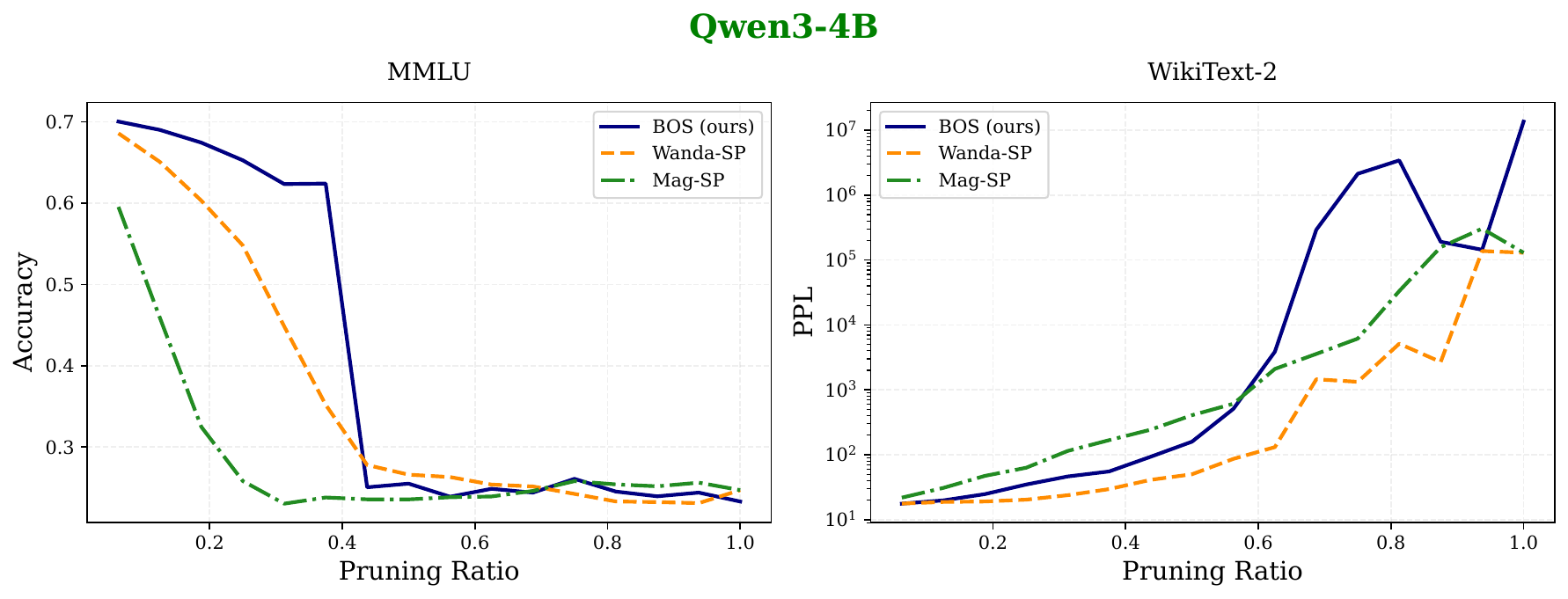}
    \caption{Performance comparison of component-level pruning methods on Gemma-3-4B, Qwen3-4B, and Llama-3.1-8B. We report MMLU 5-shot accuracy and WikiText-2 perplexity as a function of the pruning ratio. Our method consistently demonstrates stronger accuracy robustness across all models over the full range of pruning ratios compared to the baselines.}
    \label{fig:sequential_pruning_results}
\end{figure*}

\begin{table*}[p]
\centering
\small
\resizebox{\linewidth}{!}{
\begin{tabular}{c | c | c | c | cccccccc | c}
\toprule
\textbf{Model} & \textbf{Strategy} & \makecell[c]{\textbf{Pruning}\\\textbf{Ratio}} & \textbf{Wikitext-2 $\downarrow$} & \textbf{ARC-c} & \textbf{ARC-e} & \textbf{BoolQ} & \textbf{HellaS.} & \textbf{MMLU} & \textbf{OBQA} & \textbf{PIQA} & \textbf{WG} & \textbf{Avg.} \\
\midrule\midrule
\multirow{7}{*}{Gemma-3-4B} & Dense & 0 & 10.211 & 0.533 & 0.816 & 0.791 & 0.580 & 0.597 & 0.352 & 0.788 & 0.725 & 0.648 \\
\cmidrule{2-13}
 & Mag & \multirow{3}{*}{0.125} & 17.983 & 0.421 & 0.754 & 0.624 & 0.451 & 0.442 & 0.324 & 0.767 & 0.559 & 0.543 \\
 & Wanda &  & 13.532 & 0.470 & 0.786 & 0.724 & 0.508 & 0.540 & 0.318 & 0.771 & 0.635 & 0.594 \\
 & \textbf{Ours (\texttt{<BOS>})} &  & \textbf{11.271} & \textbf{0.522} & \textbf{0.813} & \textbf{0.770} & \textbf{0.571} & \textbf{0.594} & \textbf{0.348} & \textbf{0.788} & \textbf{0.718} & \textbf{0.641} \\
\cmidrule{2-13}
 & Mag & \multirow{3}{*}{0.25} & 37.125 & 0.223 & 0.714 & 0.592 & 0.365 & 0.255 & 0.272 & 0.742 & 0.543 & 0.463 \\
 & Wanda &  & 33.109 & 0.241 & 0.689 & 0.616 & 0.341 & 0.300 & 0.246 & 0.742 & 0.537 & 0.464 \\
 & \textbf{Ours (\texttt{<BOS>})} &  & \textbf{16.028} & \textbf{0.494} & \textbf{0.777} & \textbf{0.765} & \textbf{0.548} & \textbf{0.580} & \textbf{0.328} & \textbf{0.781} & \textbf{0.720} & \textbf{0.624} \\
\midrule
\multirow{7}{*}{Llama-3.1-8B} & Dense & 0 & 7.713 & 0.550 & 0.822 & 0.829 & 0.618 & 0.653 & 0.336 & 0.791 & 0.781 & 0.673 \\
\cmidrule{2-13}
 & Mag & \multirow{3}{*}{0.125} & 15.408 & 0.384 & 0.773 & 0.580 & 0.445 & 0.248 & 0.326 & 0.762 & 0.591 & 0.514 \\
 & Wanda &  & \textbf{8.501} & 0.533 & 0.812 & 0.812 & 0.559 & 0.585 & \textbf{0.346} & 0.788 & 0.743 & 0.647 \\
 & \textbf{Ours (\texttt{<BOS>})} &  & 11.249 & \textbf{0.559} & \textbf{0.825} & \textbf{0.823} & \textbf{0.611} & \textbf{0.634} & 0.340 & \textbf{0.789} & \textbf{0.770} & \textbf{0.669} \\
\cmidrule{2-13}
 & Mag & \multirow{3}{*}{0.25} & 315.209 & 0.225 & 0.612 & 0.414 & 0.321 & 0.229 & 0.220 & 0.695 & 0.507 & 0.403 \\
 & Wanda &  & \textbf{10.751} & 0.410 & 0.796 & 0.760 & 0.487 & 0.344 & \textbf{0.330} & \textbf{0.779} & 0.668 & 0.572 \\
 & \textbf{Ours (\texttt{<BOS>})} &  & 20.550 & \textbf{0.521} & \textbf{0.810} & \textbf{0.787} & \textbf{0.589} & \textbf{0.524} & 0.316 & 0.778 & \textbf{0.722} & \textbf{0.631} \\
\midrule
\multirow{7}{*}{Qwen3-4B} & Dense & 0 & 17.311 & 0.584 & 0.804 & 0.851 & 0.532 & 0.701 & 0.294 & 0.753 & 0.672 & 0.649 \\
\cmidrule{2-13}
 & Mag & \multirow{3}{*}{0.125} & 30.915 & 0.386 & 0.662 & 0.569 & 0.401 & 0.459 & 0.248 & 0.708 & 0.585 & 0.502 \\
 & Wanda &  & \textbf{18.485} & 0.542 & 0.760 & 0.804 & 0.513 & 0.653 & \textbf{0.306} & 0.734 & 0.637 & 0.618 \\
 & \textbf{Ours (\texttt{<BOS>})} &  & 19.867 & \textbf{0.568} & \textbf{0.794} & \textbf{0.817} & \textbf{0.534} & \textbf{0.690} & 0.302 & \textbf{0.755} & \textbf{0.685} & \textbf{0.643} \\
\cmidrule{2-13}
 & Mag & \multirow{3}{*}{0.25} & 63.242 & 0.334 & 0.547 & 0.453 & 0.353 & 0.258 & 0.210 & 0.652 & 0.511 & 0.415 \\
 & Wanda &  & \textbf{20.515} & 0.486 & \textbf{0.760} & 0.668 & 0.450 & 0.550 & \textbf{0.290} & \textbf{0.729} & 0.590 & 0.565 \\
 & \textbf{Ours (\texttt{<BOS>})} &  & 34.869 & \textbf{0.534} & 0.741 & \textbf{0.747} & \textbf{0.494} & \textbf{0.653} & 0.246 & \textbf{0.729} & \textbf{0.672} & \textbf{0.602} \\
\bottomrule
\end{tabular}
}
\caption{Downstream performance under component-level pruning at ratios of 0.125 and 0.25. We compare Mag, Wanda, and our \texttt{<BOS>}-based criterion across Gemma-3-4B, Llama-3.1-8B, and Qwen3-4B, reporting Wikitext-2 perplexity ($\downarrow$) and the average score over eight zero/few-shot benchmarks ($\uparrow$).}
\label{tab:downstream_combined}
\end{table*}

\paragraph{Functional Stability} We observed that these attention signatures tend to remain input-invariant. Regardless of changes in the input sequence, individual heads appear to maintain their specific patterns. This stability suggests that attention behaviors could be interpreted as structural traits inherent to the pre-trained architecture, rather than purely dynamic responses to specific semantic content. While these observations provide meaningful insights, a more rigorous validation remains for future work to confirm the extent of this inherent stability.

\section{Robustness to Model Pruning}

\label{sec:robust_pruning_anaylsis}

To evaluate our method, we perform component-level pruning on Gemma-3-4B, Qwen3-4B, and Llama-3.1-8B, and test on MMLU (5-shot) and WikiText-2. Across all three model families, the same trend appears: performance degrades more gracefully under our criterion than under baseline pruning rules. This margin is especially clear in moderate pruning regimes, where accuracy remains close to each dense baseline. Figure~\ref{fig:sequential_pruning_results} shows that our approach is notably more robust on MMLU than strong baselines (Wanda-SP, Mag-SP).

Table~\ref{tab:downstream_combined} reports downstream results for component-level pruning. At pruning ratios of 0.125 and 0.25 across WikiText-2 and eight MCQA benchmarks, our \texttt{<BOS>}-based method achieves the best average retention in most settings.

Although \BOS perplexity rises more sharply beyond a 0.5 pruning ratio, it remains competitive in the practically useful low-to-moderate range (\(\leq 0.5\)). Given that heavier pruning often yields non-functional models, this result supports \BOS as an effective strategy for real-world compression.

\begin{table*}[p]
\centering
\small
\resizebox{\linewidth}{!}{
\begin{tabular}{l l c p{6.2cm} c l}
\toprule
\textbf{Category} & \textbf{Dataset} & \textbf{Year} & \textbf{Description} & \textbf{Shots} & \textbf{License} \\
\midrule\midrule
\multirow[c]{1}{*}{Perplexity} & WikiText-2 & 2017 & Language modeling/perplexity evaluation on Wikipedia text passages. & 0-shot & CC BY-SA 3.0 \\
\midrule
\multirow[c]{8}{*}{MCQA} & ARC-Easy & 2018 & Grade-school science multiple-choice questions (easy split). & 0-shot & CC BY-SA 4.0 \\
 & BoolQ & 2019 & Yes/no reading comprehension based on short passages. & 0-shot & CC BY-SA 3.0 \\
 & OpenBookQA & 2018 & Science multiple-choice reasoning with open-book style facts. & 0-shot & Apache 2.0 \\
 & PIQA & 2020 & Physical commonsense reasoning for everyday situations. & 0-shot & Apache 2.0 \\
 & Winogrande & 2019 & Pronoun/coreference disambiguation under commonsense constraints. & 5-shot & Apache 2.0 \\
 & MMLU & 2021 & Multi-domain knowledge and reasoning across academic/professional subjects. & 5-shot & MIT \\
 & HellaSwag & 2019 & Commonsense sentence completion and grounded continuation selection. & 10-shot & MIT \\
 & ARC-Challenge & 2018 & More difficult split of grade-school science QA. & 25-shot & CC BY-SA 4.0 \\
\midrule
\multirow[c]{3}{*}{Generation} & HumanEval & 2021 & Code generation from docstrings, evaluated by unit tests. & 0-shot & MIT \\
 & TriviaQA & 2017 & Open-domain question answering using evidence documents. & 5-shot & Apache 2.0 \\
 & GSM8K & 2021 & Multi-step grade-school math word-problem solving. & 8-shot & MIT \\
\midrule
\multirow[c]{3}{*}{\makecell[c]{Long Context\\and Retrieval}} & HotpotQA & 2018 & Multi-hop question answering requiring evidence integration across documents. & 0-shot & CC BY-SA 4.0 \\
 & QASPER & 2021 & Question answering over scientific papers (long-context comprehension). & 0-shot & CC BY 4.0 \\
 & GovReport & 2021 & Long-document government report summarization. & 0-shot & CC BY 4.0 \\
\bottomrule
\end{tabular}
}
\caption{Summary of datasets used in our pruning experiments. For long-context evaluation, we follow the LongBench configuration and report results on HotpotQA, QASPER, and GovReport.}
\label{tab:dataset_summary}
\end{table*}

\begin{figure*}[t]
    \centering
    \includegraphics[width=1.0\textwidth]{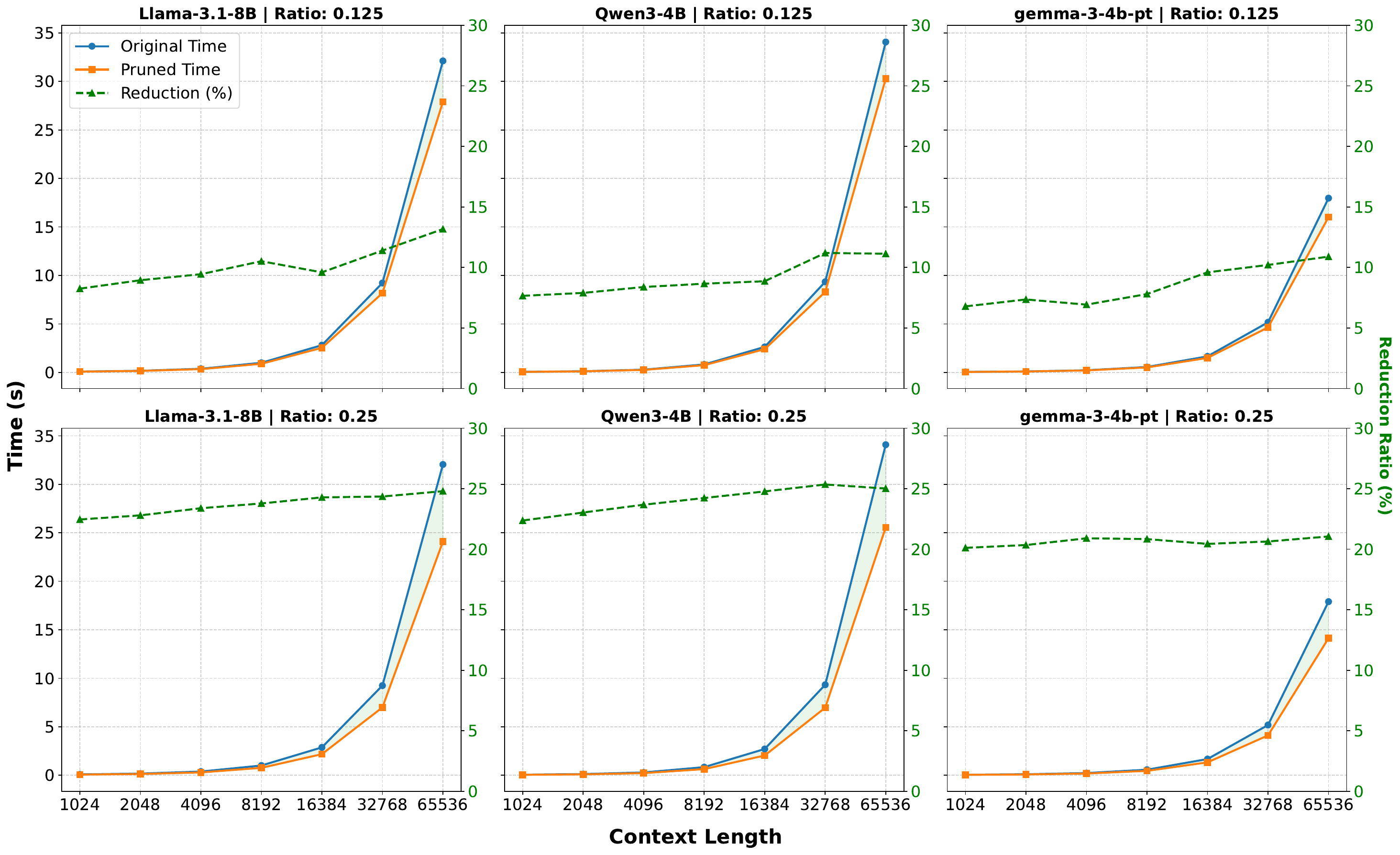}
    \caption{Wall-clock inference latency (seconds) of original and pruned models under layer-level pruning across sequence lengths from 1,024 to 65,536 tokens. Results are reported for Llama-3.1-8B, Qwen3-4B, and Gemma-3-4B at pruning ratios of 0.125 and 0.25. The proposed method consistently reduces latency across all settings, yielding about 7--13\% speedup at 0.125 pruning and 20--25\% speedup at 0.25 pruning.}
    \label{fig:latency_comparison_results}
\end{figure*}

\section{Dataset Summary for Models and Benchmarks}
\label{sec:dataset_summary}

Table~\ref{tab:dataset_summary} summarizes the benchmark suite used in our experiments, including dataset name, release year, task description, and shot setting. The protocol covers diverse capability axes---language modeling, commonsense and scientific reasoning, coding, mathematical reasoning, multi-hop retrieval, and long-document understanding---to evaluate pruning robustness across domains. Most benchmarks are evaluated in 0-shot to measure retained capability after pruning, while selected reasoning-intensive tasks (e.g., MMLU, ARC-Challenge, GSM8K, and HellaSwag) use few-shot settings following standard LLM practice. This breadth supports the consistency of our pruning observations across evaluation formats. For long-context evaluation, we follow LongBench and report HotpotQA, QASPER, and GovReport, covering both short- and long-context scenarios.

\section{Efficiency Analysis under Layer-level Pruning}
\label{sec:efficiency_analysis_layer_pruning}

\paragraph{Inference Latency.}
Figure~\ref{fig:latency_comparison_results} compares wall-clock inference latency between original and pruned models under layer-level pruning across sequence lengths from 1,024 to 65,536 tokens. We evaluate Llama-3.1-8B, Qwen3-4B, and Gemma-3-4B at pruning ratios of 0.125 and 0.25. Across all models and sequence lengths, our approach consistently reduces inference time, delivering approximately 7--13\% speedup at a pruning ratio of 0.125 and 20--25\% speedup at 0.25. These results indicate that the proposed layer-level pruning strategy provides robust and scalable latency gains, especially in long-context inference settings where computational overhead is most substantial.

\paragraph{Parameter Efficiency.}
Table~\ref{tab:parameter_efficiency_layer_pruning} summarizes parameter efficiency under layer-level pruning by reporting the total number of parameters (in billions) before and after pruning for Llama-3.1-8B, Qwen3-4B, and Gemma-3-4B. Our method consistently reduces parameter counts across all models, with reductions of approximately 9--11\% at a pruning ratio of 0.125 and 18--22\% at a pruning ratio of 0.25. These results confirm that layer-level pruning yields substantial model compression while preserving a consistent reduction trend across different model families.

\begin{table}[t]
\centering
\small
\resizebox{\linewidth}{!}{
\begin{tabular}{l c c c c}
\toprule
\textbf{Model} & \textbf{Pruning Ratio} & \textbf{Original (B)} & \textbf{Pruned (B)} & \textbf{Reduction (\%)} \\
\midrule\midrule
Llama-3.1-8B & 0.125 & 8.03 & 7.16 & 10.86 \\
 & 0.25 & 8.03 & 6.29 & 21.73 \\
\midrule
Qwen3-4B & 0.125 & 4.02 & 3.62 & 10.04 \\
 & 0.25 & 4.02 & 3.11 & 22.58 \\
\midrule
Gemma-3-4B & 0.125 & 4.30 & 3.92 & 8.78 \\
 & 0.25 & 4.30 & 3.55 & 17.56 \\
\bottomrule
\end{tabular}
}
\caption{Parameter efficiency of our method under layer-level pruning. We report the total number of parameters (in billions) before and after pruning for Llama-3.1-8B, Qwen3-4B, and Gemma-3-4B. Our method achieves parameter reductions of approximately 9--11\% at a pruning ratio of 0.125 and 18--22\% at a pruning ratio of 0.25.}
\label{tab:parameter_efficiency_layer_pruning}
\end{table}

\section{Extended IoU Analysis}
\label{sec:extended_iou_overlap}

To further characterize what our criterion removes, Table~\ref{tab:iou_head_overlap_appendix} reports pairwise IoU overlap between pruned head sets from \texttt{<BOS>}-based pruning, Wanda-SP, and Mag-SP. A consistent pattern emerges across all model families and sparsity settings: overlap between our method and Mag-SP remains low, while overlap with Wanda-SP is moderate. This indicates that our method is not a simple proxy for weight magnitude and only partially aligns with activation-based saliency. In other words, \texttt{<BOS>} sink scores identify a structurally different subset of redundant heads that conventional criteria do not fully capture. Even when sparsity increases from 0.125 to 0.25 and overlap values rise overall, this relative ordering is preserved, reinforcing that our method captures complementary redundancy signals rather than reproducing baseline selections.

\begin{table}[t]
\centering
\small
\resizebox{\linewidth}{!}{
\begin{tabular}{l c c c c}
\toprule
\textbf{Model} & \textbf{Pruning Ratio} & \textbf{BOS--Wanda} & \textbf{BOS--Mag} & \textbf{Wanda--Mag} \\
\midrule\midrule
Gemma-3-4B & 0.125 & 0.2188 & 0.1250 & 0.3438 \\
 & 0.25  & 0.3750 & 0.1354 & 0.3125 \\
\midrule
Gemma-3-12B & 0.125 & 0.3043 & 0.1159 & 0.2246 \\
 & 0.25  & 0.3992 & 0.1708 & 0.2741 \\
\midrule
Llama-3.1-8B & 0.125 & 0.3803 & 0.0810 & 0.1384 \\
 & 0.25  & 0.4679 & 0.1128 & 0.1913 \\
\midrule
Qwen3-4B & 0.125 & 0.4445 & 0.1045 & 0.1297 \\
 & 0.25  & 0.4851 & 0.1433 & 0.1755 \\
\bottomrule
\end{tabular}
}
\caption{Pairwise IoU overlap of pruned attention-head sets across \BOS-based pruning, Wanda-SP, and Mag-SP under component-level pruning.}
\label{tab:iou_head_overlap_appendix}
\end{table}

\section{Implementation Details of Wanda-SP and Mag-SP}
\label{sec:wandasp_details}

Wanda-SP and Mag-SP are the structured-pruning surrogates of Wanda and magnitude pruning introduced
as baselines in FLAP \citep[Appendix B.1]{an2023fluctuation}. Because structured pruning must remove
an entire output channel (equivalently, an entire head) rather than individual weights, the
per-weight Wanda saliency is aggregated along the row dimension of the weight matrix:
\begin{equation}
S_{i,j}^{\ell} = \left( \sum_{i=1}^{\text{row}} \left| W_{ij}^{\ell} \right| \right)
                 \cdot \left\| X_{j}^{\ell} \right\|_{2}
\label{eq:wandasp}
\end{equation}
where $|\cdot|$ denotes the elementwise absolute value and $\|X_j^{\ell}\|_2$ is the $\ell_2$ norm of
the $j$-th input feature aggregated over $N \times L$ calibration tokens. Mag-SP uses the same
aggregation with the activation term removed. Channels are ranked by $S$ and the lowest-scoring ones
are removed. Because the FLAP reference implementation assumes multi-head attention, we adapted the
score to Grouped-Query Attention by computing it per query head and removing a key/value group only
when all query heads sharing it are selected; all baselines were re-implemented under this common
framework.

\paragraph{Divergence from the original Wanda results.}
Our Wanda-SP results show substantially larger degradation than the near-lossless numbers reported
for Wanda \citep{sun2023simple} at comparable sparsity. This gap is expected, for four reasons.

\textbf{Granularity.} Unstructured Wanda prunes individual weights under a per-output-row
comparison group: every row retains its most salient weights, so the row space of each projection is
preserved and the head keeps operating, only more sparsely. Wanda-SP deletes whole heads and
channels, removing entire subspaces of the residual-stream update. The same saliency signal is thus
applied to a strictly coarser and more destructive decision.

\textbf{Sparsity levels are not comparable across the two settings.} The commonly cited Wanda
operating point (50\% unstructured sparsity) zeroes half of the parameters while keeping every head
and channel alive. Our 12.5--25\% structured ratios remove far fewer parameters but eliminate
complete architectural units, which is the more aggressive intervention in functional terms.

\textbf{No compensation.} FLAP couples structured removal with a bias-compensation term that
restores the mean output of the pruned channels; Wanda-SP as a baseline does not include this, and we
deliberately apply no post-pruning rescaling or layer-norm recalibration to any method---including
ours---so that the reported numbers reflect selection quality alone.

\textbf{Model family.} The original Wanda evaluation is centred on LLaMA-1/2, whereas we
evaluate Gemma-3, Llama-3.1, and Qwen3, all of which use GQA and therefore carry a markedly higher
functional load per head. Comparable drops for Wanda-SP and Mag-SP on generation tasks are also
reported by HARP \citep{liu2025high}, consistent with our measurements.

Wanda-SP should therefore not be read as a reproduction of unstructured Wanda; it is the closest
structured analogue of the same criterion, and it is included so that magnitude- and
activation-based selection can be compared against the \BOS sink score under an identical removal
operator.

\section{Granularity of the Pruning Unit}
\label{sec:granularity}

Section~\ref{sec:pruning_method} removes full transformer blocks even though the \BOS sink score is
computed from attention alone. To evaluate this choice, we fix the criterion and the parameter budget
and vary only the unit that is removed: Attention-only removes attention sublayers ranked by
the layer-wise \BOS score while leaving every FFN intact, and Attention+MLP removes whole
blocks under the identical ranking.

Because attention accounts for a minority of the parameters in Gemma-3-4B, a 12.5\% parameter budget
requires removing 23 of the 34 attention sublayers under Attention-only, against 4 blocks under
Attention+MLP. Table~\ref{tab:granularity_ablation} reports the outcome: Attention-only yields a
WikiText-2 perplexity of $2.5\times10^{6}$ and an MCQA average of 0.331, close to the chance level of
these eight benchmarks, whereas Attention+MLP yields 20.02 and 0.589. Meeting the budget through
attention alone therefore requires removing attention from most of the network, including the layers
the criterion ranks as least prunable, whereas block-level removal draws its 4 blocks from the
high-scoring deeper layers (Figure~\ref{fig:layer_ablation_trend}).

\begin{table}[t]
\centering
\small
\resizebox{\linewidth}{!}{
\begin{tabular}{l | c | c | c | c}
\toprule
\textbf{Target} & \makecell[c]{\textbf{Pruning}\\\textbf{Ratio}} & \makecell[c]{\textbf{\# Layers}\\\textbf{Pruned}} & \textbf{Wikitext-2 $\downarrow$} & \textbf{Avg. $\uparrow$} \\
\midrule\midrule
Attention only & 0.125 & 23 & 2528008.47 & 0.331 \\
Attention + MLP & 0.125 & 4 & $\mathbf{20.02}$ & $\mathbf{0.589}$ \\
\bottomrule
\end{tabular}
}
\caption{Effect of the pruning unit on Gemma-3-4B at a 0.125 parameter budget. Both rows use the same layer-wise \BOS sink ranking and differ only in what is removed. \textbf{Avg.} is the mean over the eight MCQA benchmarks of Table~\ref{tab:layer_pruning_results}.}
\label{tab:granularity_ablation}
\end{table}

\section{Perplexity across Model Scales and Pruning Ratios}
\label{sec:ppl_scale_ratio}

The perplexity cost of structured pruning depends jointly on the pruning ratio and on model size. To
characterise this, we extend the layer-level comparison of Table~\ref{tab:layer_pruning_results} to
pruning ratios from 0.0625 to 0.5 and add Gemma-3-27B, evaluated under the same WikiText-2 protocol
as the other models.

Table~\ref{tab:ppl_scale_ratio} reports the results. At 0.0625 all three models stay within roughly
two to four points of their dense baselines (12.62 against 10.21 on Gemma-3-4B, 14.42 against 10.18
on Gemma-3-12B, and 9.26 against 7.45 on Gemma-3-27B). The ratio at which perplexity leaves a usable
range increases with model size: at 0.125 the pruned-to-dense ratio is 1.96 on Gemma-3-4B, 1.70 on
Gemma-3-12B, and 1.61 on Gemma-3-27B, and Gemma-3-27B still reaches 22.41 at a 0.25 ratio. Beyond
0.25 every strategy degrades sharply, but the gap between them widens rather than closing: at 0.375
on Gemma-3-12B, ShortGPT reaches $6.84\times10^{12}$ while our criterion stays at
$2.33\times10^{3}$. Across the twelve pruned settings, our criterion attains the lowest perplexity in
ten, counting the tie with ShortGPT at 0.0625 on Gemma-3-12B.

\begin{table}[t]
\centering
\small
\resizebox{\linewidth}{!}{
\begin{tabular}{l | c | c | cccc}
\toprule
\textbf{Model} & \makecell[c]{\textbf{Pruning}\\\textbf{Ratio}} & \textbf{Dense} & \textbf{Ours (\texttt{<BOS>})} & \textbf{ShortGPT} & \textbf{Bottom Up} & \textbf{Top Down} \\
\midrule\midrule
\multirow{3}{*}{Gemma-3-4B}
 & 0      & 10.21 & --- & --- & --- & --- \\
\cmidrule{2-7}
 & 0.0625 & --- & $\mathbf{12.62}$ & \underline{12.90} & 16.52 & 19.35 \\
 & 0.125  & --- & $\mathbf{20.02}$ & \underline{20.82} & 50{,}643.64 & 45.68 \\
\midrule
\multirow{6}{*}{Gemma-3-12B}
 & 0      & 10.18 & --- & --- & --- & --- \\
\cmidrule{2-7}
 & 0.0625 & --- & $\mathbf{14.42}$ & $\mathbf{14.42}$ & \underline{16.00} & 20.51 \\
 & 0.125  & --- & $\mathbf{17.27}$ & \underline{34.53} & 147.38 & 46.85 \\
 & 0.25   & --- & $\mathbf{63.98}$ & $3.06\times10^{8}$ & 3{,}968.23 & \underline{191.38} \\
 & 0.375  & --- & \underline{2{,}327.06} & $6.84\times10^{12}$ & $3.72\times10^{7}$ & $\mathbf{896.58}$ \\
 & 0.5    & --- & $\mathbf{5{,}292.32}$ & $1.40\times10^{10}$ & $1.95\times10^{11}$ & \underline{312{,}570.92} \\
\midrule
\multirow{6}{*}{Gemma-3-27B}
 & 0      & 7.45 & --- & --- & --- & --- \\
\cmidrule{2-7}
 & 0.0625 & --- & \underline{9.26} & 9.50 & $\mathbf{8.43}$ & 17.52 \\
 & 0.125  & --- & $\mathbf{12.01}$ & \underline{13.73} & 107.19 & 197.59 \\
 & 0.25   & --- & $\mathbf{22.41}$ & \underline{24.96} & 273.96 & 953.51 \\
 & 0.375  & --- & $\mathbf{86.62}$ & 46{,}875.04 & 48{,}826.19 & \underline{22{,}201.11} \\
 & 0.5    & --- & $\mathbf{1{,}038.98}$ & \underline{14{,}295.21} & $1.13\times10^{14}$ & $8.08\times10^{6}$ \\
\bottomrule
\end{tabular}
}
\caption{WikiText-2 perplexity ($\downarrow$) under layer-level pruning across model scales and pruning ratios. Bold and underline denote the best and second-best result per row.}
\label{tab:ppl_scale_ratio}
\end{table}

\begin{figure*}[p]
    \centering
    \includegraphics[height=0.95\textheight]{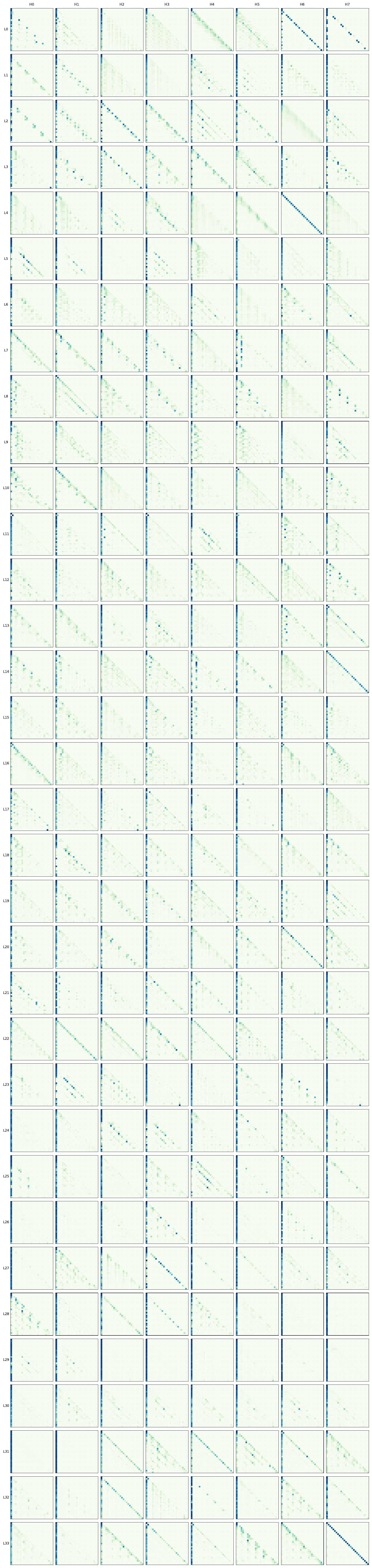}
    \caption{Full Attention Maps for Gemma-3-4B.}
    \label{fig:full_attention_maps_gemma}
\end{figure*}

\begin{figure*}[p]
    \centering
    \includegraphics[width=1.0\textwidth]{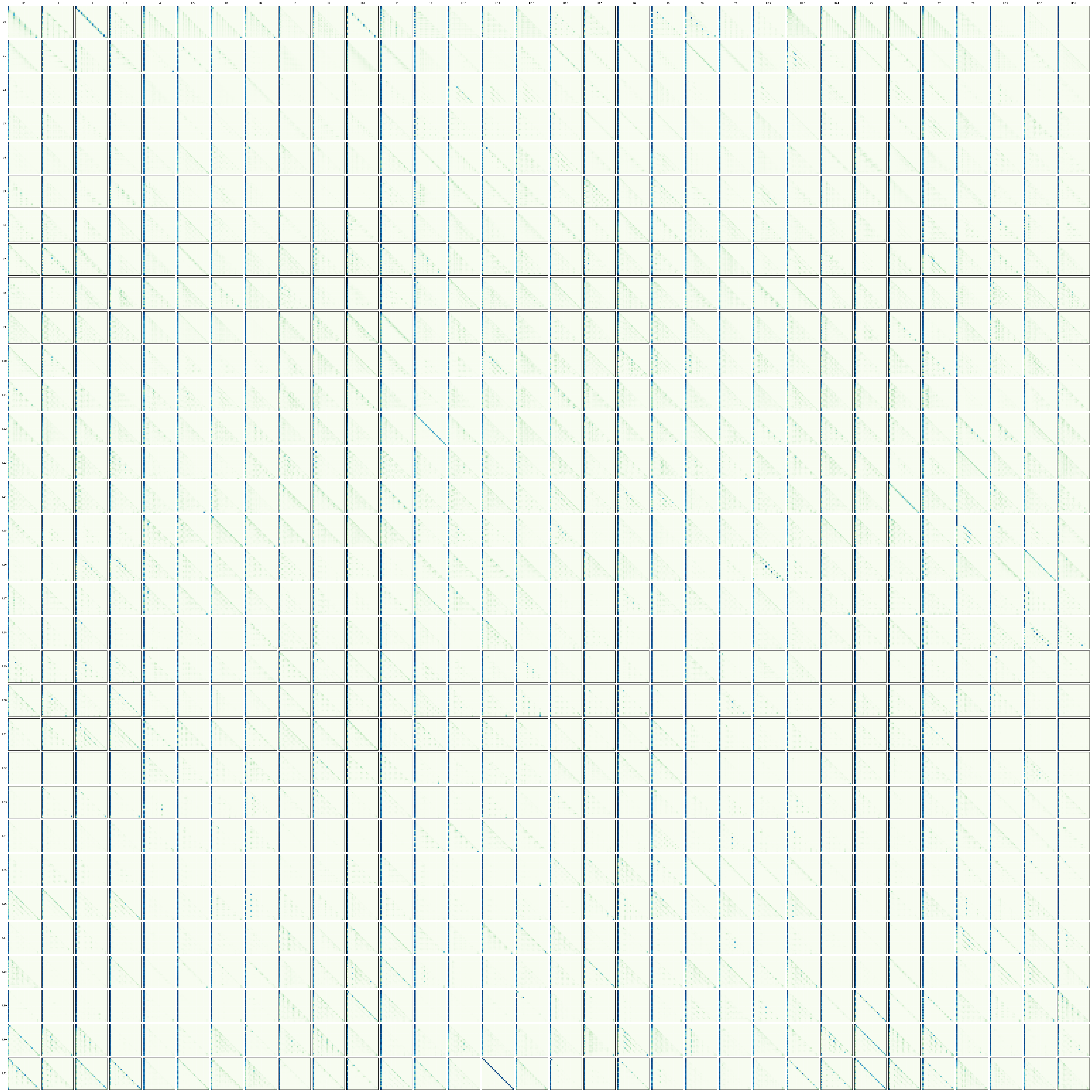}
    \caption{Full Attention Maps for Llama-3.1-8B.}
    \label{fig:full_attention_maps_llama}
\end{figure*}

\begin{figure*}[p]
    \centering
    \includegraphics[width=1.0\textwidth]{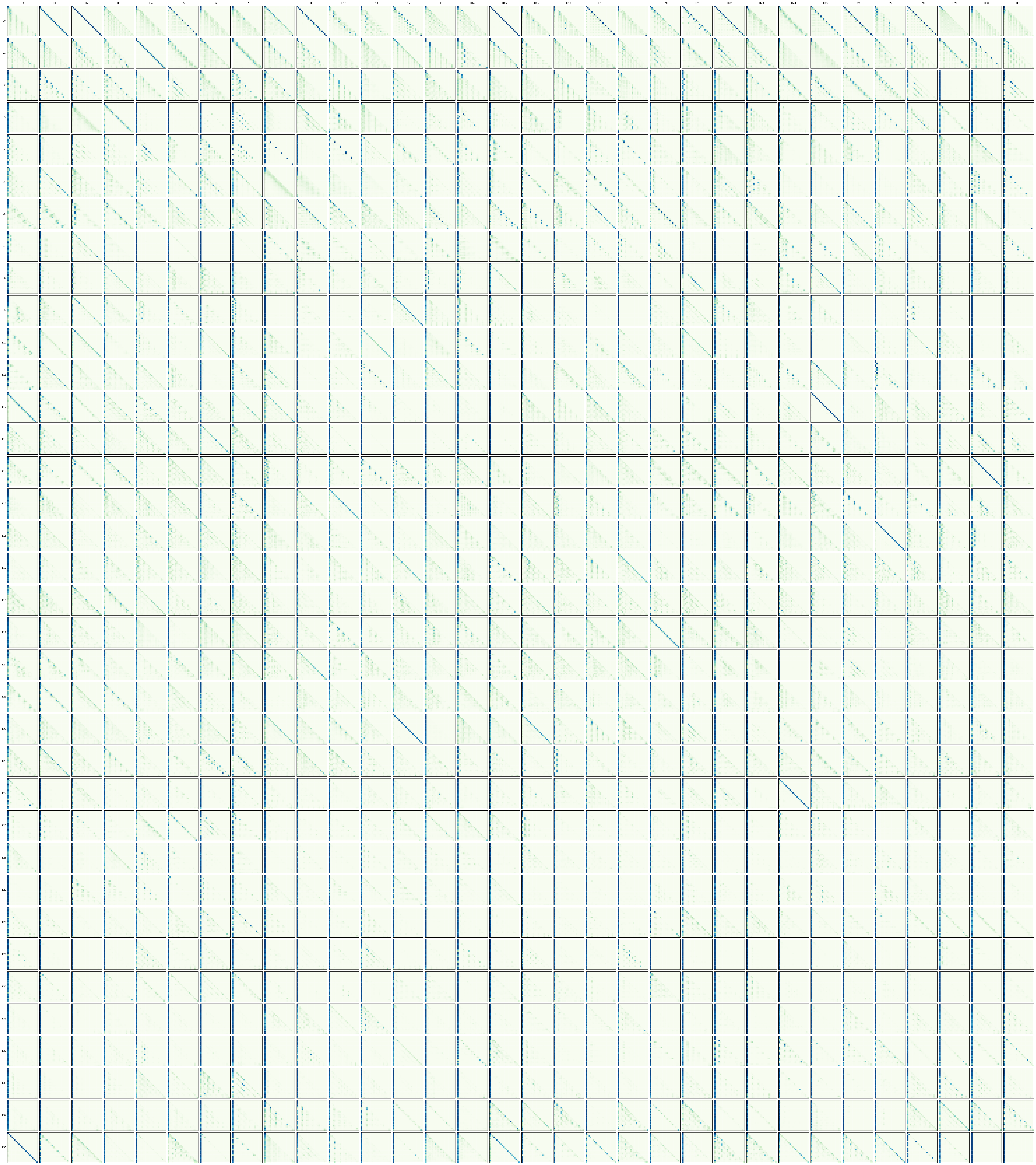}
    \caption{Full Attention Maps for Qwen3-4B.}
    \label{fig:full_attention_maps_qwen}
\end{figure*}

\end{document}